# Implementation and Evaluation of a Multivariate Abstraction-Based, Interval-Based Dynamic Time-Warping Method as a Similarity Measure for Longitudinal Medical Records

**Matan Lion and Yuval Shahar**

Medical Informatics Research Center,

Department of Software and Information Systems Engineering,

Ben-Gurion University of the Negev,

Beer-Sheva,

Israel

Matan.lion@gmail.com, yashahar@bgu.ac.il

**Corresponding Author:**
Yuval Shahar
Department of Software and Information Systems Engineering
Ben-Gurion University of the Negev
Beer-Sheva 8410501,
Israel
**yshahar@bgu.ac.il**

# Abstract

**Objectives**: A common prerequisite for tasks such as classification, prediction, clustering and retrieval of longitudinal medical records is a clinically meaningful similarity measure that considers both [multiple] variable (concept) values and their time. Currently, most similarity measures focus on raw, time-stamped data as these are stored in a medical record. However, clinicians think in terms of clinically meaningful temporal abstractions, such as "decreasing renal functions", enabling them to ignore minor time and value variations and focus on similarities among the clinical trajectories of different patients. Our objective was to define an abstraction- and interval-based methodology for matching longitudinal, multivariate medical records, and rigorously assess its value, versus the option of using just the raw, time-stamped data.

**Methods**: We have developed a new methodology for determination of the relative distance between a pair of longitudinal records, by extending the known *dynamic time warping* (DTW) method into an *interval-based dynamic time warping* (iDTW) methodology. The iDTW methodology includes (A): A three-steps *interval-based representation* (iRep) method: [1] abstracting the raw, time-stamped data of the longitudinal records into clinically meaningful interval-based abstractions, using a domain-specific knowledge base, [2] scoping the period of comparison of the records, [3] creating from the intervals a symbolic time series, by partitioning them into a predetermined temporal granularity; (B) An *interval-based matching* (iMatch) method to match each relevant pair of multivariate longitudinal records, each represented as multiple series of short symbolic intervals in the determined temporal granularity, using a modified DTW version.

**Evaluation**: Three classification or prediction tasks were defined: (1) classifying 161 records of oncology patients as having had autologous versus allogenic bone-marrow transplantation; (2) classifying the longitudinal records of 125 hepatitis patients as having B or C hepatitis; and (3) predicting micro- or macro-albuminuria in the second year, for 151 diabetes patients who were followed for five years. The raw, time-stamped, multivariate data within each medical record, for one, two, or three concepts out of four or five concepts judged as relevant in each medical domain, were abstracted into clinically meaningful intervals using the Knowledge-Based Temporal-Abstraction method, using previously acquired knowledge. We focused on two temporal-abstraction types: (1) *State* abstractions, which discretize a concept's raw value into a predetermined range (e.g., LOW or HIGH Hemoglobin); and (2) *Gradient* abstractions, which indicate the trend of the concept's value (e.g., INCREASING, DECREASING Hemoglobin value). We created all of the combinations of either *uni-dimensional* (State *or* Gradient) or *multi-dimensional* (State *and* Gradient) abstractions, of all of the concepts used. Classification of a record was determined by using a majority of the *k-Nearest-Neighbors* (*KNN*) of the given record, *k* ranging over the odd numbers (to break ties) from 1 to $\sqrt{N}$, *N* being the size of the training set. We have experimented with all possible configurations of the parameters that our method uses. Overall, a total of 75,936 experiments were performed: 33,600 in the Oncology domain, 28,800 in the Hepatitis domain, and 13,536 in the Diabetes domain. Each experiment involved the performance of a 10-fold Cross Validation to compute the mean performance of a particular iDTW method-configuration set of settings, for a specific subset of one, two, or three concepts out of all of the domain-specific concepts relevant to the classification or prediction task on which the experiment focuses. We measured for each such experimental combination the *Area Under the Curve* (AUC) and the optimal Specificity/Sensitivity ratio using Youden's Index. We then aggregated the experiments by the types of unidimensional or multidimensional abstractions used in them (including the use of only raw concepts as a special case); for example, two state abstractions of different concepts, and one gradient abstraction of a third concept. We compared the mean AUC when using each such feature *representation*, or combination of abstractions, across all possible method-setting configurations, to the mean AUC when using as a feature representation, for the same task, only raw concepts, also across all possible method-setting configurations. Finally, we applied a paired t-test, to determine whether the mean difference between the accuracy of each temporal-abstraction representation, across all concept and configuration combinations, and the respective raw-concept combinations, across all concept subset and configuration combinations, is significant (P<0.05).

**Results**: The mean performance of the classification and prediction tasks when using, as a feature representation, the various temporal-abstraction combinations, was significantly higher than that performance when using only raw data. Furthermore, in each domain and task, there existed at least one representation using interval-based abstractions whose use led, on average (over all concept subset combinations and method configurations) to a significantly better performance than the use of only subsets of the raw time-stamped data. In seven of nine combinations of domain type (out of three) and number of concepts used (one, two, or three), the variance of the AUCs (for all representations and configurations) was considerably higher across all raw-concept subsets, compared to all abstract combinations. Increasing the number of features used by the matching task enhanced performance. Using multi-dimensional abstractions of the same concept further enhanced the performance. When using only raw data, increasing the number of neighbors monotonically increased the mean performance (over all concept combinations and method configurations) until reaching an optimal saddle-point around $\sqrt{N}$; when using abstractions, however, optimal mean performance was often reached after matching only five nearest neighbors.

**Conclusions**: Using multivariate and multidimensional interval-based, abstraction-based similarity measures is feasible, and consistently and significantly improved the mean classification and prediction performance in time-oriented domains, using DTW-inspired methods, compared to the use of only raw, time-stamped data. It also made the KNN classification more effective. Nevertheless, although the *mean* performance for the abstract representations was higher than the *mean* performance when using only raw-data concepts, the actual optimal classification performance in each domain and task depends on the choice of the specific raw or abstract concepts used as features.

**Keywords:**
Dynamic Time Warping, Temporal Abstraction, Interval-Based Similarity Measure, Knowledge Representation, Temporal Data Mining, Classification

# 1. Introduction

## 1.1. **The Need for Matching Longitudinal Patient Health Records**

Multiple tasks, across many different organizations, involve the continuous collection and maintenance of large quantities of longitudinal data. Examples include web server logs that store a user's behavior, motion sensors that document and maintain a pedestrian's movements over time and space, possibly captured also by street cameras, and patient health records that store the patient's medical history and measurements, as reported by various medical laboratory tests, or possibly by a care provider. To extract succinct but relevant knowledge from such a quantity of data, efficient and effective data-mining processes are needed.

The importance of providing automated support is particularly relevant in the medical domain, in which most tasks usually involve measurement and capture of numerous time-stamped longitudinal patient data. Mining longitudinal patient records has a significant potential for facilitating health care, including support to tasks such as *Classification* (e.g., determining a patient's diagnosis, given all of the patient's symptoms and additional data), *Prediction* (e.g., predicting a patient's outcome given the disease course up to a present point), *Clustering* (e.g., grouping patients who share a common course of the disease process), and *Retrieval* (e.g., retrieve the patients most similar to a given patient, based on their records, for determination of future treatment of the given patient).

Another example of a specific need, which is closely related to retrieval, is the task of finding similar patients in social networks that connect patients. For instance, *PatientsLikeMe* is a patient networking site that enables people with various disorders to share their experiences with others who share the same disease or condition. Using this capability, the users can compare their experiences to those of other patients and provide them with better understanding of their treatment options [Brubaker et al., 2010]. Another interesting social network is *CureTogether*, which utilizes methods such as Collaborative Filtering to identify patients in similar situations [Swan, 2009].

### 1.1.1. Similarity Measures for Matching Longitudinal Patient Records

A closer look at the above tasks reveals a common denominator related to the characteristics of all of the methods underlying these tasks: A *distance measure* between the observed objects can be used to solve each of the tasks. Such a measure is also referred to as a (*dis*)*similarity function*. In order to apply such a measure, it is crucial to understand the underlying definition of such a function. To begin with, similar objects are defined as having traits or characteristics in common. Similarity can be defined as the closeness of the appearance of one object to another object. Subsequently, the similarity function represents a real-valued function that quantifies the similarity between two objects. In the clinical domain, the objects are often patients, represented by the patient's longitudinal electronic health record; the similarity between two time-oriented records might be composed of common events, measurements, treatments, and diagnoses of the observed patients over time; the similarity function could be calculated by the overall distance between these components.

Therefore, it is clear that there is a need for a similarity function that is appropriate for each relevant clinical domain, and that whose application will satisfy a quality measure defined by comparison to a predefined gold standard (e.g., some objective classification or prediction accuracy).

Point-based time-series similarity measures are the most common similarity measure; their numerous variations and flavors had been reviewed in detail by Wang et al. [2012]. The point-based measures are mostly divided into two subcategories: (1) The *Lock-Step Measures*; and (2) The *Elastic Measures*.

*Lock step measures* are a set of methods (e.g., the Euclidean-Distance) designed to compare each *i*th point in one time series to the corresponding *i*th point in the other series in a linear manner. Although the Lock step measures, such as the Euclidian distance, are hard to compete with, especially when the length of the time series increases [Serra and Arcos, 2012], they often fail to reflect the underlying meaning of similar sequences in the user's perspective [Park and Kim, 2006]. *Elastic measures*, on the other hand, differ from the Lock-Step measures by allowing a comparison cardinality other than one-to-one, and are defined as a set of methods aiming to compare time series with similar patterns, sampled at different paces or time, by stretching (or compressing) the time dimension while estimating the distance between the observed series.

*Dynamic Time Warping* (DTW) [Brendt and Clifford, 1994], which originated from the speech recognition community, a domain in which the time series are notoriously complex and noisy [Sun and Giles, 2001], became a popular elastic distance measure and a solution for time-series matching problems in a variety of domains, including medicine [Tormene et al., 2009], bioinformatics [Aach and Church, 2001], security [Kovács-Vajna, 2000], and more. In contrast to the lock step measures, the DTW algorithm is designed to cope with the comparison of sequences with different lengths, and its major virtue lies in its ability (at least, theoretically) to identify similar patterns within two time series, even if the similar patterns are located at different periods in time, within each of the compared series.

Nevertheless, the classic DTW algorithm is often unsuitable for many real-life domains, as such those that include a set of *multiple* features sampled *episodically* at different points in time, since the original algorithm is limited to only *one* variable, and assumes a *complete* data set, without any gaps. Furthermore, as was previously already noted explicitly by Kotsifakos et al. [2013], almost no study has referred to the matching of *interval-based* time-series, although such a matching is a typical need of real-life applications, including the medical domain [Kostakis et al., 2011]. Moreover, these similarity measures typically focus on analyzing only *raw*, time-stamped data as they are stored in the database. However, human clinicians are often highly effective in thinking about meaningful subsets of patients, because they abstract the raw, time-stamped data into clinically meaningful *temporal abstractions*, which are clinically meaningful concepts derived from the raw, time-stamped data, in a context-sensitive fashion. Examples include a *State* abstraction of the Hemoglobin value, such as "*three months of severe anemia in the context of a young female*", and a *Gradient* abstraction of multiple concepts related to kidney functions, such as "*a decrease in renal functions during the five months following chemotherapy*". Temporal abstractions enable human experts (and, we think, also automated systems) to ignore minor variations in specific values and time stamps, while noticing the similarity among the clinical trajectories of different patients

### 1.1.2. The Objectives of the Current Study

To suggest a comprehensive solution that can support a widely differing set of tasks, and in particular, classification, prediction, clustering, and retrieval, we have developed a new, interval-based version of DTW, *iDTW*. The iDTW method is a generic methodology for determining the similarity between two or more interval-based, usually knowledge-driven, longitudinal multivariate electronic health records. Based on this methodology, we implemented the *iDTW Framework,* a computational framework for the classification, prediction, and clustering of multivariate time-stamped entities, which greatly extends the DTW concept.

The main goal of the current study is to [1] present an overview of the iDTW methodology, and then [2] to evaluate, using that methodology, the value, for classification and prediction purposes, of determining the distance between two longitudinal medical records by matching multivariate, knowledge-based time intervals, which represent a knowledge-based interpretation (*abstraction*) of the data within each record, versus determining that distance by matching only the time-stamped *raw* data of the two records.

Note that using abstractions might well lose crucial information, and might not equal the use of the full raw data set.

Other than presenting the core principles of iDTW, the evaluation objectives of the current study can be further specified as follows:

1 ) **Evaluation** of the **effectiveness of the use of interval-based abstractions for similarity matching**, by assessing the expected performance of three classification and prediction tasks, one in each of three different clinical domains (hepatitis, diabetes, and oncology), using only the symbolic, interval-based abstractions of the raw, time-stamped data, versus the performance of these tasks when using the full raw time-stamped data set.

2 ) **Assessment** of the added value of using ***multi-dimensional*** temporal abstractions of the *same* concept (e.g., both State and Gradient abstractions of the Hemoglobin concept) for similarity matching, for the expected performance of the classification and prediction tasks, as well as multivariate, multiple-concept abstractions.

i.e., several different temporal abstractions of the *same* concept, in addition to the use of *uni-dimensional* abstractions, i.e., a single abstraction for each concept.

**1.1.3. Our Main Contributions**

In the current study we are presenting several different contributions:

1. An interval-based, temporal-abstraction-based methodology for determining the distance between different multivariate time series: The iDTW methodology;

2. A rigorous evaluation of the expected value of the use of abstractions of time-oriented data, versus the use of the raw time-stamped data;

3. A precise assessment of the expected value of multidimensional abstractions of the same concept, in addition to univariate abstractions of different concepts, as additional classification and prediction features;

4. An assessment of the expected number of neighbors needed when using KNN to determine the correct classification or prediction, in the case of interval-based temporal abstractions, versus when using raw time-stamped data.

# 2. Background

## 2.1. Temporal Abstraction

*Temporal Abstraction* (TA) is the task of interpreting time-stamped raw data into a set of time intervals, over which hold more abstract concepts. As a result, it provides a more powerful, concise, and integrated description of a collection of time-stamped raw data [Shahar and Combi, 1998].

Temporal abstraction techniques usually include a discretization module that discretizes raw data into a small number of values. This discretization can be performed either by automatic data driven discretization techniques [Azulay et al., 2007] or by using knowledge acquired from a domain expert.

In general, automatic data-driven discretization techniques are divided into two categories: (1) *unsupervised discretization* techniques (e.g., Equal Width Discretization, Equal Frequency Discretization), some being more time-oriented, such as SAX [Lin, Keogh et al., 2007] or PERSIST [Mörchen and Ultsch, 2005]); and (2) *Supervised discretization* techniques (e.g., ID3 and C4.5, or the interval-based TD4C method [Moskovitch and Shahar, 2015c]), which rely on the existence of target-class information. In particular, the *temporal discretization for classification* (TD4C) method, and even three slightly different versions of it, has been demonstrated to have several computational and performance advantages compared to the unsupervised methods, while not requiring any explicit domain knowledge beyond the existence of the target classes [Moskovitch and Shahar, 2015c].

However, since they were usually designed to analyze static data, and assume that no domain knowledge is available, and often focus only on discretization of states, but not of other abstractions, such as gradients or trends, automatic data-driven discretization techniques typically do not sufficiently consider the time dimension or existing domain knowledge. Thus, in the current study, we used a domain-specific knowledge-based technique (the knowledge being acquired from domain experts) that is specific to the temporal-abstraction task, based on the *Knowledge-Based Temporal-Abstraction* (KBTA) theory [Shahar, 1997] for the representation of time-series vectors as an input for enhanced time-oriented, knowledge-specific matching methods.

Several types of interval-based, context-dependent abstractions can be derived by KBTA: (1) *State* Abstraction – The abstracted, context-dependent, state of a concept over some time interval (e.g., HIGH blood pressure in the context of a woman's pregnancy); (2) *Gradient* Abstraction – The direction (INREASING, DECREASING, SAME) in which the data are changing in a specific context (e.g., increasing blood pressure while pregnant); (3) *Rate* Abstraction – The speed in which the measured concept is changing within a specific context (e.g., SLOW increase of BP while pregnant); and (4) *Pattern* Abstraction – The combined time and value relationships and constraints that exist between different intervals composed of different types (e.g., decreasing liver functions *during* the administration of a Statin-type medication).

The KBTA method is composed of five computational mechanisms that operate in parallel to generate the temporal abstractions. Details can be found elsewhere [Shahar, 1997]. However, typical key mechanisms include *temporal-context induction*, which generates, given an external event (such as a medication administration) or a temporal abstraction (such as "moderate anemia") interval-based *temporal contexts*, within which other abstraction mechanisms operate in a context-sensitive manner; *contemporaneous abstraction*, which generates State abstractions using a context-sensitive classification function; and *temporal interpolation*, which exploits several domain-specific knowledge types. The temporal-interpolation knowledge aspects include *local persistence function*, *global persistence function*, and the *significant variation functions*. The *local persistence functions* represent the local persistence of the truth of value of a concept instance, given a single concept-value point or interval, before ("Good-Before") or after ("Good-After") the temporal scope of the concept instance. The *Global persistence function* on the other hand, bridges the gap between two different instances of a concept's value, using a function that returns the maximal time gap that still enables joining the instances into a single abstraction that is believed to be true. The *significant variation* function is used to induce the Gradient-abstractions values (INCREASING, DECREASING, SAME), by defining the sufficient change between two sequential instances' values, usually by a single threshold [absolute value or percentage].

The KBTA method has been shown to produce useful abstractions that are similar to those produced by domain experts in multiple clinical domains, such as oncology, diabetes and children's growth assessment [Shahar and Musen, 1993, 1996], and that are useful for multiple tasks in other domains, such as for detecting malware threats in the cyber security domain [Shabtai, Shahar, and Elovici, 2006a, 2006b, 2010]. KBTA-generated temporal abstractions have also been successfully used to visualize and explore the longitudinal records of individual patients' data [Shahar et al., 2006; Martins et al., 2008], and to visualize, explore, and find associations among large numbers of multiple patients' data [Klimov et al., 2009, 2010a, 2010b]. The same temporal abstractions have been demonstrated to be the basis for automatically generating, in an accurate and highly complete manner, summaries in free text of the multivariate data of longitudinal medical records. Examples include automatically creating discharge summaries of intensive care unit (ICU) patient records [Goldstein et al., 2016, 2017].

Frequent combinations of temporal abstractions (i.e., combinations that exists above some proportion threshold of the patient population) have been shown to create temporal patterns that can be effectively used as features for classification and prediction [Moskovitch and Shahar, 2015a, 2015b, 2015c]. Fortunately, the frequent temporal patterns consistently repeat within different subsets of a patient population, for each clinical domain [Shknevsky, Shahar, and Moskovitch, 2017]. The distribution of such frequent temporal patterns within a patient's record has been shown to be a major useful meta-feature for prediction, such as for predicting sepsis in the ICU [Sheetit et al., 2019].

Figure 1 displays a typical set of raw time-stamped data and derived temporal abstractions for a patent who has had a bone-marrow transplantation procedure and was monitored for a year.

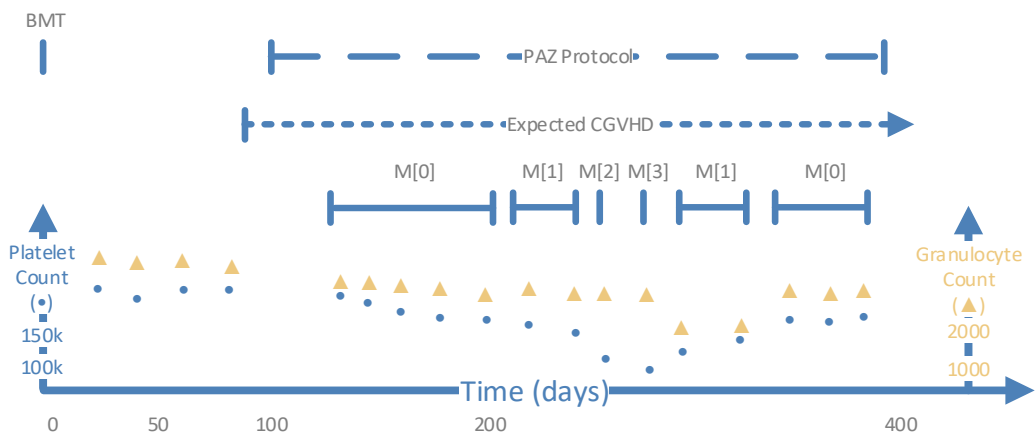

**Figure 1** – Temporal abstraction of platelet and granulocyte values during abstraction of a prednisone/azathioprine (PAZ) clinical protocol for treating patients who have chronic graft-versus-host disease (CGVHD). The time line starts with bone-marrow transplantation (BMT) – an input external event. The platelet and granulocyte count concepts and the PAZ and BMT external

events are typical inputs. The abstraction and context intervals are typically part of the output. • = platelet counts; ∆ = granulocyte counts; dashed interval = event; shaded arrow = open context interval; ├─┤ = closed abstracted interval; M[*n*] = myelotoxicity {Bone-marrow toxicity} grade *n*.

## 3. Similarity Measures for Time Series

Various methods have been proposed to cope with the different issues, which in turn produced several works that summarize and compare them [Wang et al., 2012; Warren, 2005].

*Lock step measures* are a set of methods designed to compare each *i*th point in one time series to the corresponding *i*th point in the other series in a linear manner. Euclidean distance and its variants is one of the first generic techniques used to compare time series [Faloutsos et al., 1994]. It is characterized by an intuitive approach to implement linear computation time complexity relative to the series' lengths, a competitive scalability, and independence of external variances [Wang et al., 2012]. Although it is not targeted at comparing sequences of different lengths, it is possible to re-sample one sequence along the length of the other [Serra and Arcos, 2012]. The Euclidian distance, also referred to as $L_2$, is part of the $L_p - Norms$ family. Other members of this family include the Manhattan distance ($L_1$) and the Maximum norm ($L_\infty$). The Pearson correlation coefficient is a statistical Lock Step measurement that represents the degree of the linear relationship between two variables. Previous studies conducted in the bioinformatics domain, including by Ewing [1999] and by Häne et al. [1993] used Pearson's correlation as a statistical measurement of similarity between DNA fingerprints, and provided an example of the use of this measurement in the longitudinal data field.

Although the Lock step measures, such as the Euclidian distance, are hard to compete with, especially when the length of the time series increases [Serra and Arcos, 2012], they often fail to reflect the underlying meaning of similar sequences in the user's perspective [Park and Kim, 2006]. The *elastic measures* differ from the Lock-Step measures by allowing a cardinality that is different from one-to-one, and are defined as a set of methods aiming to compare time series with similar patterns, sampled at different paces or time.

The *Longest Common Subsequence* (LCSS) model aimed at easing the Lock Step measures' restrictions and at finding the longest subsequence common to both a query and a reference sequence. Vlachos et al. [2002] extended the above user perception by controlling the flexibility allowed between items both in the time and in the space dimensions. Different approaches, measuring the distance by penalizing the gaps between the matched subsequences are presented by the Edit Distance with Real Penalty (ERP) [Chen and Ng, 2004] and by the Edit Distance on Real Sequences (EDR) [Chen et al., 2005] measures.

### 3.1. Dynamic Time Warping

The *Dynamic Time Warping* (DTW) algorithm [Brendt and Clifford, 1994] is a popular solution for time series problems encountered in a variety of domains, including medicine [Tormene et al., 2009], bioinformatics [Aach and Church, 2001], security [Kovács-Vajna, 2000], and others.

DTW is designed to calculate the optimal path between two time series of different lengths, using a dynamic-programming methodology, while minimizing the effect on accuracy in comparison with equal-length time series comparisons [Ratanamahatana & Keogh, 2004].

By duplicating the matching of previous elements and enabling local time shifting to minimize the distance between two time series, also known as the *DTW Warping Distance*, DTW is able to identify similar patterns within two time series, even if located at different periods in time, within each of the compared series. The *Warping Distance* is defined by the summation of the optimized path on the Cost Matrix R($m, n$), in which $dist(X_i, Y_j)$ is the local distance between the two instances $X_i$ and $Y_j$ of the matched time series $X$ and $Y$, of length $m$ and $n$, respectively.

In the case of *univariate* time-series, the Euclidian distance is considered as an acceptable alternative for the local distance. In the case of *multivariate* time-series, or perhaps when domain-knowledge is available, a suitable local function that considers all variables should be considered. However, the individual-variable $L_p - Norms$, such as the Euclidian distance, can be combined, as suggested by [Giorgino et al., 2007; Rath & Manmatha, 2002], who used the sum of the normalized squared Euclidian distance of each dimension in the multivariate space to calculate the dissimilarity within the local distance.

Restriction of the matching period between two time-series can be used to reduce run time and to increase the accuracy in the task to be solved. Park et al. [2001] proposed the Prefix-querying approach based on a sliding window technique, using a constant-duration prefix for the query pattern. Extending the Prefix-query method, Park and Kim [2006] employed an $l_1$ norm instead of the common $l_\infty$ norm as the distance function. Tormene et al. [2009] proposed the *Open-Ended-DTW* (*OE-DTW*) algorithm, which compares a query sequence to the prefix of each of the reference sequences that are closest to it, for the problem of motion recognition. Similarly, Forestier et al. [2015] used DTW partial matching for predicting the next step a surgeon is about to perform during a surgery, based on the on-going surgical processes and a prerecorded set of complete surgeries.

Local constraining of the warping path for a limited distance $r$ from the main diagonal, representing the optimal alignment, is quite standard, although in theory the constraint can be adjusted over the entire range, from the tightest window to the most relaxed variation (i.e., no constraint). The vast majority of researchers have used a window size of 10% [of the total series' duration] width for the global constraint [Aach and Church, 2001; Sakoe and Chiba, 1978], a setting that seems to derive from historical reasoning rather than from an empirical reference or special property of the constraint [Ratanamahatana and Keogh, 2004]. Other studies [Chaovalitwongse, et al., 2007; Kostakis et al., 2011] implemented the traditional DTW algorithm while imposing no restriction on the warping path, possibly under the assumption that perhaps a wider warping window improves the matching performance.

Other optimization techniques can be applied to DTW, including setting a (1) *Slope limitation* – restricting the warping path to a somewhat moderate gradient, in order to prevent unrealistic correspondence between a short segment to a long segment; and an (2) *Index-based approaches* – a set of methods that aim to prune a substantial amount of sequences that could not possibly be the best match [Keogh and Ratanamahatana, 2004], including the lower-bound (LB) approach introduced by Park et al. [2001], which utilizes four distinct features from the sequence, but can only be applied to similar length time-series.

Other variations of DTW have focused on using only a simplified distance measure, such as the use of only the shape of the two curves defined by the matched time series $X$ and $Y$, represented by the difference between the local derivatives of the two curves as the sole distance measure, as demonstrated successfully through the *derivative-based DTW* algorithm, by Keogh and Pazzani [2001]. As we shall see when we describe our methodology, in a sense, some of our work on designing the iDTW methodology in the current study can be seen as extending the Keogh and Pazzani derivative-based approach to general temporal abstractions (not just gradients), and to the use of not just the direction, but also the actual values, and to the use of longer, interval-based durations (and not just point-based local derivatives).

**3.1.1. Interval-Based Dynamic Time Warping**

Classic DTW refers to point-based time series matching, as is the case for most time-series matching measures, whereas many daily tasks include also interval-based time-series.

Kostakis et al. [2011] suggested the Vector-based DTW Distance as an extension to the classic DTW by matching, using the $L_1$ distance, the distance between two *E-Sequences*, i.e., an ordered set of event intervals $S = \{S_1, ..., S_n\}$, such that $S_i = (E_i, t_{start}^i, t_{end}^i)$ refers to the start and end times of the event interval, and $E_i \epsilon \sigma$, is taken from a finite alphabet of the events' labels. Kostakis et then mapped the e-sequences to a sequence of event vectors, such that each event vector indicates which and how many intervals with the same symbol are active at each time point, where a time point is defined as a period in which the state of the active intervals does not change. Although simple to implement, the vector-based approach does not explicitly take into account the temporal relations (e.g., *overlap*) that may occur between the interval-based events, or (unlike our temporal-abstraction approach) the internal semantics of the events (e.g., their possible range of values, and the meaning of the differences between two values).

As we show in the next section, the iDTW approach we suggest considers, among other details, the time intervals of each and every entity as first-class citizens for matching, including both their value and duration.

# 4. Methods

## 4.1. iDTW: A Symbolic Multivariate Interval-Based Dynamic Time-Warping Method

To assess the distance between two or more longitudinal time series through the use of interval-based, symbolic *(abstraction-based)* representations, and using our previous work on temporal abstraction in general, and on knowledge-based temporal abstraction in particular, we have developed a *symbolic multivariate interval-based dynamic time-warping method* (iDTW). Based on the iDTW methodology, we have implemented the iDTW Framework, a computational framework for the classification and prediction of multiple entities.

The iDTW methodology is composed of two main subtasks - an *interval-based representation* task, the *iRep* task, and an *interval-based matching* task, the *iMatch* task.

The first subtask, *iRep*, transforms the raw data into a domain-specific meaningful representation within a two-dimensional space (i.e., a set of interval-based temporal abstractions) that can be used for the matching task. This transformation can be performed either via automatic data-driven discretization techniques that do not have any knowledge regarding the domain, as we have explained in detail in Section 2.1, or, as we do in the current study, via knowledge-based techniques, using the *Knowledge-Based Temporal Abstraction* (KBTA) method [Shahar, 1997] and through a domain-specific temporal-abstraction context-sensitive knowledge base acquired from a domain expert. The temporal abstractions are further processed via temporal segmentation, scaling and normalization, aggregation, and interpolation procedures to form a homogenous two-dimensional multivariate time-series, i.e., a series of intervals over which various abstract concepts hold.

The second subtask, *iMatch*, focuses on the application of an extended version of a DTW-based distance measure that exploits the multivariate iRep time-series representation and the domain-specific knowledge for the matching task.

Figure 2 presents an overall view of the iDTW methodology.

We shall now proceed to describe each of the iDTW method's sub-tasks and the computational mechanisms performing them.

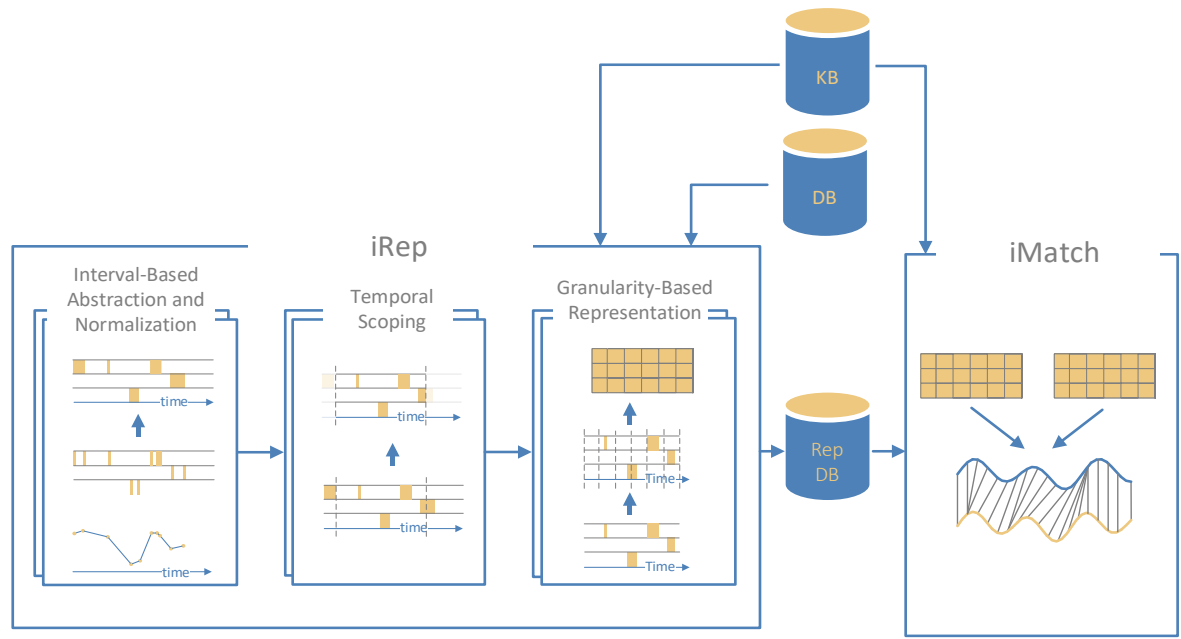

**Figure 2:** An overall view of the iDTW methodology. The process includes two main tasks: (1) *iRep*, to transform the raw data, includes three steps; and (2) *iMatch*, to match the transformed series. KB = knowledge base; DB = database. All iRep procedures use the domain-specific KB. The iRep input is a point-based DB, its output is an interval-based representation (Rep) DB.

## 4.2. The iRep Method

The iRep subtask of the iDTW methodology includes three main procedures:

(1) Creating a normalized knowledge-based temporal abstraction of the raw data;

(2) Temporal scoping;

(3) Granularity-based representation.

We shall now discuss the three iRep procedures in detail.

### 4.2.1 Generating a Normalized Knowledge-Based Temporal-Abstraction of the Raw Data

The first phase of the iRep subtask transforms the input time-stamped raw data into a series of time intervals over which hold meaningful abstractions of the input data.

The abstraction and normalization of the time-stamped raw data transforms these data into a higher level representation, which typically embodies a context-sensitive interpretation. In other applications, such a representation was found to be more comprehensible to humans and to support faster and more accurate decision making, especially when using a domain-specific *Knowledge Base* (KB) applied through the KBTA method [Shahar and Musen, 1996; Martins et al., 1998; Klimov et al, 2009, 2010b]. It was also found to be useful when frequently-discovered temporal patterns are formed by discovering all of the temporal relations among the interval-based knowledge-based abstractions, and used as machine-learning features for classification and prediction [Moskovitch and Shahar, 2015b] and for repeatedly discovering the same temporal patterns in different groups of patients within the same medical domain [Shknvesky et al., 2017].

Several different temporal abstraction types are defined by the KBTA ontology. In our current implementation of iDTW and in our evaluation study, we had focused on the generation and use of the following two complementary abstraction types: (1) the *State* abstraction, which refers to the classification (or computational transformation) of the value of one or more concepts (e.g., LOW, HIGH abstraction for the Hemoglobin concept); and (2) the *Gradient* abstraction, which represents the sign of the derivative of the concept's value (e.g., INCREASING, DECREASING, SAME (STABLE), abstractions of the value of the Hemoglobin parameter).

To support both the computation of the distance between two values of raw data and the computation of the distance between two knowledge-based symbolic abstractions of these data during the matching process, a uniform representation of the temporal abstractions was defined, in which each concept value was mapped into a numeric value. Its application depended on the type of each abstract concept, as defined by the KBTA ontology: Boolean (False and True values), Ordered Symbolic, which are measured on an Ordinal scale (e.g., LOW, NORMAL, HIGH), and Numeric (e.g., 0 to 100), which are measured on an Interval scale (having an equal distance between every two consecutive values of the scale, such as temperatures measured on the Celsius or Fahrenheit scale) or on a Ratio scale with a 0 point (such as weight).

Since the numeric output type in the case of abstractions (e.g., a Body Mass Index, computed from Height and Weight) is a trivial case, it requires no further mapping, retaining its actual value. However, to eliminate the potential bias of representing different concepts using different values ranges, all abstracted concepts' values were scaled using a Min-Max Scaling approach, to the interval [0, 1]. The output types that require a special treatment are Boolean and Ordered (ordinal-scale) Symbolic. In both cases the symbolic values of the abstract concept were converted into a set of natural integers (starting from 1), following the original order as defined in the knowledge base. Thus, both the values of ordinal symbolic concepts, such as the white blood-cell (WBC) State abstraction, i.e., Low, Normal, and High, or of the WBC Gradient abstraction, i.e., Decreasing, Same, and Increasing, were converted into the values 1, 2, and 3. Boolean False and True values were converted into the values 1 and 2. Then, a Min-Max Scaling normalization, assuming equal distances between the ordinal symbolic values (i.e., an interval scale) was performed, reducing all abstract values into numbers within the [0..1] range.

Finally, for purposes of the evaluation, so as to enable a fare comparison to the standard DTW algorithm, in the case of using raw time-stamped values without any abstraction, the raw numeric concepts values were first standardized to the population's parameters using a Z-score normalization, and then scaled to the [0, 1] interval.

Following the nomenclature of Kostakis et al. [2011], regarding interval-based events, albeit with somewhat different semantics, we refer to the output of the first phase of the iRep task, temporal abstraction and normalization, as a set of domain-specific, interval-based *Multivariate E-Sequences*, derived from the values of the concepts' set $C$ for the entities set $M$, and denoted as $\mathbf{M}^{C(m)}$. More concretely, for entity (patient) $m$, $\mathbf{M}^{C(m)} = \{U^{c_1(m)}, U^{c_2(m)}, \ldots, U^{c_n(m)}\}$, where $U^{c_i(m)}$ is a univariate e-sequence, representing the data of concept $c_i \in C, 1 \leq i \geq |C|$, such that event interval $S_j^{c_i(m)}$ refers to the $j$th interval of concept $c_i$ for entity $m$.

### 4.2.2 Performing a Temporal Scoping

The second iRep phase constrains the temporal scope of the E-Sequences that were generated by the first iRep phase.

Scoping the temporal space supports a more focused and relevant matching of the time-series in the context of the domain in hand, by setting a restriction on the temporal span to be matched, limiting it to a period relative to a predetermined significant event in the domain, instead of matching the entire longitudinal record.

*4.2.2.1 Catering for Relative and Absolute Timelines*

Based on the domain-knowledge in hand, we set the timeline to be used for the matching task to be either a *Relative Timeline* or an *Absolute Timeline*.

The *Absolute Timeline* anchors the temporal events based on the calendric time of the samples, whereas in the Relative Timeline, the events are located in a timeline that is relative to a certain significant *reference event*. In such a setting, the definition of time is modified by defining the zero-time of the timeline as the time in which the reference event occurred, and shifting the other events relatively in time.

The *Relative Timeline* [Shahar et al., 2006; Shahar and Combi 1998] is set by identifying clinically significant milestones, that are typically aspects of a reference event, to which we refer as *Reference-Point Aspect Functions*, in the domain's temporal-abstraction ontology (e.g., the End-time of the Bone-Marrow transplantation procedure event). Thus, a milestone includes a *Reference Concept*, i.e., a time-stamped clinical concept of some type; a *Reference point aspect function*, to denote the temporal aspect of the reference concept; and a *Reference-Point Selection Function*, for selecting a specific instance of the chosen Reference Concept if more than one exists. We have defined three reference selection functions: (1) *First* – the first sampled instance of the reference concept type within the longitudinal record's timeline; (2) *Last* – the last sampled instance of the reference concept type; and (3) *N* – the Nth ordered instance of the reference concept type.

*4.2.2.2 Full and Partial Scope Matching*

To support a dynamic matching of a restricted time periods, relevant to the problem to be solved, the iRep scoping process supports two temporal scoping modes – a *Full-scope Matching* and a *Partial-scope Matching*.

Full-scope Matching typically uses an absolute timeline, in which a matching-scope absolute start time and end time must be provided, denoted $abs\_start$ and $abs\_end$ in accordance. Alternatively, when a Relative Timeline is used, the following are required (1) a reference point aspect function (*RPAF*) (e.g., RPAF = First occurrence of a BMT event); (2) the *Before Period* (*BP*); (3) and the *After Period* (*AP*) – in which the RPAF will be mapped to the appropriate absolute point in time in the timeline of entity $E_i$, RPAF(*i*); (e.g., $BP = -1\ Month$ and $AP = +2\ Years$, meaning a period spanning the one month before and up to two years after the RPAF for patient *p*).

The overall *Temporal Matching Scope* (*TMS*) for entity $E_i$, within which the matching should occur, $TMS_i$, is defined as follows, depending on the type of timeline to be used:

$$TMS_i = \begin{cases} \{abs\_start, abs\_end\} & \text{Absolute Timeline} \\ \{\text{RPAF}(i) - BP, \text{RPAF}(i) + AP\} & \text{Relative Timeline} \end{cases}$$

Finally, the output of the scoping and timeline setting is a *Restricted-Scope Multivariate E-Sequence* $M_{(TMS_i)}^{C(i)} = \{U_{(TMS_i)}^{c_1(i)}, U_{(TMS_i)}^{c_2(i)}, \ldots, U_{(TMS_i)}^{c_n(i)}\}$ such that each $U_{(TMS_i)}^{c_k(i)}$ is defined as a restricted univariate e-sequence based on the same boundaries, in which the event-interval $S_j^{c_k(i)}$ is included if the start of the interval is included within the restricted scope, the end of the interval is within the restricted scope, or, alternatively, if the restricted scope is included within the interval.

To express these constraints more formally, we include in the analysis exactly the following interval-based components:

$$S_j^{c_k(i)}.t_{start} \geq TMS_i.t_{start} \cap S_j^{c_k(i)}.t_{start} \leq TMS_i.t_{end}$$

$$OR\ S_j^{c_k(i)}.t_{end} \geq TMS_i.t_{start} \cap S_j^{c_k(i)}.t_{end} \leq TMS_i.t_{end}$$

$$OR\ S_j^{c_k(i)}.t_{start} < TMS_i.t_{start} \cap S_j^{c_k(i)}.t_{end} > TMS_i.t_{end}$$

### 4.2.3 Creating a Granularity-Based Representation

In the third and last phase of the iRep subtask, we perform a segmentation of the intervals generated and scoped in the first two phases, into a continuous time series of abstract-concept values, at a predetermined temporal granularity. Such a representation is sufficient for applying all versions of the DTW algorithm, including the multivariate matching version that we are using for the iMatch task.

To perform the temporal-granularity segmentation, we have defined a three-part process, to which we refer as a *Granularity-Based Representation (GBR) of multivariate time-series*. A GBR process is capable of supporting the input requirements of most of the time-series matching algorithms, which require a point-based representation of a given variable's values. The GBR method includes (1) a segmentation of the Restricted-Scope Multivariate E-Sequence generated by the former iREP phase (Temporal Scoping), (2) an imputation of missing (abstract) values, and (3) an aggregation of points and intervals.

The output of the GBR process is a two-dimensional matrix of the multivariate time series, at the appropriate temporal granularity, representing a value for each of the concepts, for every time granule.

The first GBR step, also denoted as the *Granularity-based Multivariate Segmentation (GMS)* method generates the required structure to support the multivariate time-series matching, performed during the iMatch task. The generated scoped intervals, all at the same required time granularity, are then further transformed in the second and third steps of the GBR process, namely, the imputation of missing (abstract) values and the aggregation of points and intervals. These two steps simply fill with values (if needed) the time granules created in the first step. The GMS method receives as input a restricted multivariate e-sequence $M_{(TMS_m)}^{C(m)}$, and a predefined required time granularity $G$, and returns a two-dimensional matrix $\overline{M}_{TMS_i}^{C(i)}$, referred to as a *Multivariate Restricted-Scope Event Table*. The matrix represents entity $E_i$ at equal time intervals (also referred as *time-granules*), for example, hours or days, within a predefined restricted period $TMS_i$ that was defined previously by a relative or absolute scope. The columns of the multivariate event table $\overline{M}_{TMS_i}^{C(i)}$ reflect the number of segments sampled in the new representation and can be defined as the number of time-granules in the restricted period $TMS_i$, i.e., $Length(TMS_i, G) = Duration(TMS_i.t_{start}, TMS_i.t_{end}, G)$, where the $Duration(X, Y, G)$ operator refers to the number of time granules of size G between two time points $X$ and $Y$. Each of the time points of the new multivariate sequence is a time point $t_j, 1 \leq j \leq length(TMS_i, G)$. (Note that each "time point" might in fact represent a period, such as a whole month, depending on the choice of the time-granule G.) The number of rows of $\overline{M}_{TMS_i}^{C(i)}$ is the number of matched concepts $c \in [1, ..., |C|]$, where $|C|$ corresponds to the number of different (abstract) concepts (e.g., the State of Hemoglobin, the Gradient of WBC count…). Thus, each column represents the set of co-occurring concepts at time point $t_j$. A time-granule of concept $x$ at time point $y$ is therefore defined as $\overline{M}_{TMS_i}^{C(i)}[x, y]$.

We also need to make sure that within each time-granule included in the *Multivariate Restricted-Scope Event Table* provided by the GMS method, for each (abstract) concept, there is only one value (e.g., High). Aggregation of multiple intervals within the same granule for each concept (e.g., each lasting several hours, within the same day) is handled by predefined representative functions, also referred to as *Delegate Functions*, inspired by Klimov et al.'s work on visualization and exploration of time-oriented data and their abstractions [Klimov et al., 2010b]. Namely, given an entity $E_i$, for each non-empty time granule $\overline{M}_{TMS_i}^{C(i)}[x, y]$, the representative function calculates a *delegate value* for the whole time-granule interval, which represents during the analysis the (single) value of this time-granule for the concept $c_x$. The delegate function is a function of the values existing within the time-granule's scope (e.g. Maximum, Minimum, Mode, Longest Duration Interval) that is specific to the concept $c_x$. The function $Duration(X)$ refers to the duration of interval $X$, namely $X.t_{end} - X.t_{start}$:

*Value-Based aggregation*: If all event intervals included in $\overline{M}_{TMS_i}^{C(i)}[x, y]$ have a similar duration, such that there is no significance to the duration of each interval, we can use as a *delegate function* that returns a single value of the whole interval, one of several standard statistical aggregation functions, such as Max, Min, Mean, Median, Mode, to determine the representative value for the time-granule. The assessment of the conditions that enable the application of the Value-based aggregation can be formalized as follows:

$$\forall k, l, Duration\left(\overline{S_y^{c_x(i)}}[k]\right) = Duration\left(\overline{S_y^{c_x(i)}}[l]\right), 1 \leq k, l \leq \left|\overline{S_y^{c_x(i)}}\right|.$$

Which statistical function best represents a single value for a time-granule that includes several values (e.g., three blood-glucose state values within the same day) is determined using the domain knowledge base, which includes a default suggested statistical aggregation function for each concept and time-granule. For numeric values, the default delegate function is the Mean; for ordinal values, the Median; for categorical non-ordinal values, the Mode.

*Duration-Based Aggregation*: Otherwise, in the case in which at least one event interval included in $\overline{M}_{TMS_i}^{C(i)}[x, y]$ has a duration that is different from the duration of some another interval, we use a more general *delegate function*. In a general delegate function, both the values and durations of the intervals are used for assessment of the granule's representative value. When using this technique, we use either (1) the *Maximal Total Time* (*MTT*), which refers to the concept value that has the maximal cumulative duration over all of the relevant intervals within the relevant time-granule aggregation time period; or (2) the *Longest Interval* (*LI*), which refers to the concept value that holds over an interval that has the longest duration during that time period.

The assessment of the conditions that enable the application of the Duration-based aggregation can be formalized as follows:

$$\exists k, l, s.t. Duration(\overline{S_y^{c_x(i)}}[k]) \neq Duration(\overline{S_y^{c_x(i)}}[l]), 1 \leq k, l \leq \left|\overline{S_y^{c_x(i)}}\right|$$

The decision of which general delegate function to use is also determined using the domain knowledge base. The default is the MTT aggregation function.

*Filling empty time granules with missing values:* Nevertheless, even after performing the previously explained steps, and in particular the KBTA-based interval-based temporal abstraction, there might still remain time-granules, within the Multivariate Restricted-Scope Event Table, of the desired granularity, within whose scope there is no value of one or more state-abstraction concepts. However, to efficiently apply a DTW procedure, and match two different vectors, we need each of the vectors to represent a continuous time series (at the same temporal granularity) without missing values.

To support the assignment of values to empty time granules $\overline{M}_{TMS_i}^{C(i)}[x,y]$, while leveraging the characteristics of the temporal intervals to be matched, iDTW uses several standard functions. In particular, in our main experiment we used all of the following three: The *Nearest Neighbor* interpolation function (duplicating the value of the nearest point), *Linear interpolation* (which projects the value of the time point onto a linear line between the values of its neighbors, considering, in computing the value, the specific time granule at which the value is needed), and *Average interpolation* (using simply the mean value of the neighbors).

We also experimented, completely separately from our main evaluation, as we briefly discuss in the Discussion Section, with a novel approach that we have defined that exploits the previous step of abstracting the data into interval-based, knowledge-based temporal abstractions, which we refer to as *Interval-Based Adjacent Proportional interpolation* (*IBAP*). Using the IBAP method, the values of state-abstraction concept within a gap of any duration that exists between two intervals (also of any duration) that have *different* state-abstraction values for that concept are generated by considering the duration of the two intervals adjacent to the empty (i.e., without an assigned concept value) interval. The IBAP procedure extends the values of the concepts of the two neighboring intervals into the empty time-interval gap in a manner proportional to their durations. Thus, the temporal duration of each value within the gap is proportional to the duration of the respective adjacent intervals on both sides of the previously empty interval, over which each value holds. A detailed description of the IBAP method and of its justification can be found in the Appendix. We briefly discuss the results of this experiment in the Discussion section.

In summary, the overall resulting output of the iRep task, and in particular, its final GBR process, is a two-dimensional matrix of the multivariate time series, at the desired temporal granularity, representing a value for each of the concepts, for every time granule. Such a representation, which is equivalent to a series of time points at the chosen temporal granularity (albeit, time points that have abstract values), is easy to use by multiple time-series analysis algorithms, and in particular, by the DTW algorithm (including our version of it, as we shall see when we describe the matching phase).

### 4.3. **The iMatch Task**

Given the input of a series of abstract concepts holding over time points at a predetermined temporal granularity, the second and final subtask of the iDTW methodology is the iMatch subtask. This phase focuses on the actual matching of the transformed time-series, based on the iRep output, using methods that are quite similar to the original DTW method, while including several algorithmic modifications and domain-specific customizations of the algorithm's parameters.

**Comparing multiple values at each point:** One key requirement in multivariate time-oriented data analysis (unlike the original examples for DTW application, such as speech-wave identification) is to support the concurrent matching of multiple values at each time point, such as multiple clinical measures. To achieve that within iDTW, having segmented the abstract, interval-based concepts into the desirable time granularity, we are applying a previously introduced modification of the classic DTW univariate matching algorithm [Chaovalitwongse et al., 2007; Giorgino et al., 2007; Kostakis et al., 2011; Rath and Manmatha, 2002]. In this modification, the local DTW distance, $d_{local}$, between time point $k$ and $l$ of the two matched sequences of entities $E_i$ and $E_j$, at each of which a vector of measurement holds, is defined as the extended Squared Euclidean Distance between the two time points, computed over all concept values. Thus, the distance between the two time series that are being compared, denoted as the vectors $V_k^i$ and $V_l^j$, is computed by summing the extended Squared Euclidean Distances over all concepts, between all of the paired points of the two compared vectors. For each pair of points, and for $|C|$ concepts, the local distance is defined as:

$$d_{local}(V_k^i, V_l^j) = \sum_{n=1}^{|C|} \left(V_k^i[n] - V_l^j[n]\right)^2$$

Assuming sufficient domain-specific knowledge, additional DTW matching parameters can adjusted to better uncover the true semantics underlying the similar groups, speed up the matching process, and also save preprocessing time.

**Constraining the time-warping window:** To avoid pathological matching, prevent irrelevant patterns from being discovered, and also to accelerate the matching process by narrowing the possible warping paths needed to be considered within the cost matrix, the Sakoe-Chiba band is usually used for constraining the warping window width [Sakoe and Chiba, 1978]. The typical values used in most studies are a window size whose duration is 10% of the overall time-series duration, or Infinity (no constraint). These are the two values we used in our main evaluation of the iDTW methodology.

We also experimented, completely separately from our main evaluation, as we briefly discuss in the Discussion Section, with a novel window-constraining approach that we have defined. Since the maximal window span of 10% of the overall time series duration seems rather arbitrary, as previously noted by several researchers [Ratanamahatana and Keogh, 2004], we have defined a new *Knowledge-Based-Band (KB-Band)*. Inspired by the Sakoe-Chiba band, the KB-Band implements a domain-specific temporal band, or window size, defined as the maximal half-life of all of the matched concepts $C$ involved in the matching task, to induce the value $r$ (the size of the band) of the Sakoe-Chiba band. A detailed description of the KB-Band approach and its justification can be found in the Appendix. We briefly discuss the results of this experiment in the Discussion section.

### 4.3.1 Unidimensional versus Multidimensional Matching

Since we enrich the raw data based on multiple temporal abstractions, we distinguish between matching a time series of data that include only a single abstraction of a concept (e.g., WBC State) and matching a time series of data that include two or more abstractions of the same concept (e.g., both the WBC State and the WBC Gradient concepts), by referring to the first as *Unidimensional* time-series matching, and to the second as *Multidimensional* time-series matching.

Moreover, both variations of unidimensional and multidimensional matching can be applied when using either only a single concept, i.e., *Univariate* time-series matching, or when applied on multiple concepts, i.e., *Multivariate* time-series matching.

In our evaluation, we have assessed the expected contribution of both multivariate and multidimensional matching to the expected performance of several classification or prediction tasks, based on the iDTW distance measure and a version of the *K-Nearest Neighbor* (KNN) algorithm, as we explain the Experimental Design section.

## 5. Evaluation

The main objective of the evaluation was to use iDTW to match time-series data in three different clinical domains, using knowledge-driven, interval-based, domain-specific interpretations of the raw numeric data and raw data, applying both univariate and multivariate time-series matching, and answer the primary research question:
 ***Does the use of interval-based temporal abstractions of the data improve the expected (mean) performance of each of several classification and prediction tasks, compared to their expected (mean) performance when using only raw data?***

Secondary research questions examined various aspects of the iDTW methodology, and in particular, (a) the expected value of multivariate and multidimensional matching, (b) the expected effect of the number of concepts abstracted (in particular, we compared uni-variate, bi-variate, and tri-variate concept matching), and (c) the mean number of records needed for a *K*-Nearest Neighbors (KNN)-based classification.

Note that *absolute classification performance was not an objective*; thus, to sharpen and elucidate the questions and the answers, and to make our results easily comparable to the literature, we focused only on the relative value of the various representations of the time-series data, and did not even include demographic data. Also, and for the same reasoning, we used the *K-Nearest Neighbors* (KNN)-based classification, which is very common when comparing different temporal data representations techniques and time-series similarity measures, as was the case in multiple previous studies [Kostakis et al., 2011; Kotsifakos et al., 2013; Ratanamahatana and Keogh, 2004; Sheetrit et al., 2019; Tormene et al., 2009; Wang et al., 2012], since KNN is a natural candidate method for a distance measure.

## 5.1. Datasets and Domain-Specific Knowledge

Three different retrospective longitudinal datasets were used for the evaluation (see Table 1); *Institutional Review Board* (IRB) permission had been given to use the data for each of the three datasets. All sets included only unidentifiable data. The datasets included (1) An oncology dataset from a major medical center that was provided for our analysis; (2) A hepatitis dataset that was previously published publicly as part of a previous KDD challenge [Ho et al., 2003]; and (3) a Type 2 Diabetes dataset of patients from a large health maintenance organization, provided to us to investigate the prediction task that we shall describe. For each data set, we focused on four or five time-stamped concepts whose temporal behavior we wished to characterize.

**Table 1** - Descriptive statistics of the three evaluation datasets. For each dataset the number of records, entities (i.e., patients), concepts, and average records per entity are presented.

| Dataset | Records | Entities | Concepts | Average Records Per Entity |
|---|---|---|---|---|
| **Oncology** | 34,228 | 161 | 5 | 213 |
| **Hepatitis** | 53,760 | 125 | 5 | 430 |
| **Diabetes** | 10,659 | 151 | 4 | 71 |

The knowledge bases included, for each of the concepts on which we had focused, the State-abstraction classification function in the relevant context, and the Gradient abstraction Significant Variation in the relevant context. Default Good-Before and Good-After persistence values were acquired as well. The relevant knowledge used to support the framework's capabilities was provided by an internist and was complemented by examining the medical literature.

The domain-specific knowledge bases of the three domains are described in the Appendix.

### 5.1.1. The Oncology Dataset

The oncology dataset is a retrospective database of more than 1000 unidentifiable laboratory-test records of bone-marrow transplantation patients. The data contains more than 200 raw concepts (e.g., White Blood Cell count [WBC], Hemoglobin count) and different recorded events (e.g., bone marrow transplantations [BMT], platelet transfusion).

*Selection criteria:* We used the records of 161 patients who have had a bone marrow transplantation (Autologous or Allogenic) procedure, and who had at least one sample for the following five hematological laboratory tests, which were considered as potentially relevant in this domain, and were sufficiently frequently measured during a period of one month after the patient's first transplantation: White Blood Cell count (WBC), Platelet Count (PLATELET), Hemoglobin value (HGB), Monocytes count (MONOS), and Neutrophilic band forms count (BANDS). (Having more than one bone marrow transplantation is not infrequent. Thus, in this evaluation we chose to concentrate on the first occurrence as a default, and focused on the main recovery period.)

The *task* was to classify the records into records of patients who went through autologous bone-marrow transplantation (110 patients) versus records of patients who have had allogenic bone-marrow transplantation (51 patients).

### 5.1.2. The Hepatitis Dataset

The hepatitis dataset is a publicly accessible dataset that contains the results of laboratory examinations of the patients with hepatitis B or C, who were admitted to Chiba University Hospital in Japan. Hepatitis A, B, and C are viral infections that affect the liver of the patient. Hepatitis B and C chronically inflame the hepatocytes, whereas hepatitis A acutely inflames it. Hepatitis B and C are especially important because they have a potential risk of developing liver cirrhosis or hepatocarcinoma. The dataset contains long time-series data on laboratory examinations of 771 patients who were examined between 1982 and 2001.

The data include administrative information such as the patient's information (Age and Date of birth), pathological classification of the disease, date of biopsy, result of biopsy (i.e., type of hepatitis), and duration of interferon therapy. Additionally, the data include the temporal records of blood tests and urinalysis. The temporal data contain the results of 983 types of examinations. The relevant knowledge used to support the evaluation was extracted from the public KDD challenge and was used in our evaluation (see Appendix).

*Selection Criteria:* We used 125 patients who went through Liver Biopsy in order to reveal the type of Hepatitis they carry that resulted in Hepatitis B or C, and also had at least one of the following five laboratory tests during a period of three months before their *last* biopsy: ALkaline Phosphatase (ALP), Direct (conjugated) BILirubin (D-BIL), Indirect (unconjugated) BILirubin (I-BIL), Total BILirubin (T-BIL), and Lactate DeHydrogenase (LDH). (T-BIL = D-BIL + I-BIL, but not all three, or at least two of them, are always reported). (Note: Patients who suffer from hepatitis may undergo several Biopsy procedures. This may occur due to a wrong diagnosis, or perhaps as a precaution. Therefore, we wished to focus on the most relevant hepatitis diagnosis of the patient.)

The *task* was to classify patients to Hepatitis B (65 patients) versus C (60 patients).

### 5.1.3. The Diabetes Dataset

The diabetes dataset included 26,833 diabetes patients of a large health maintenance organization, who have undergone various periodic laboratory tests between 2004 and 2008. The data included static information (e.g., Gender and Date of birth) and temporal records (e.g., Alkaline Phosphatase, Lipoprotein, Triglycerides, Glucose, Hemoglobin, Creatinine, Total cholesterol and Albuminuria).

*Selection criteria:* We used 151 patients who had Urine Albumin collected for 24 hours (ALBUMINURIA-U24h) during the second year after the patient's first observation, and who also had in the first year this test and at least one sample for the following 3 laboratory tests: Glycosylated Hemoglobin (HbA1c), Creatinine (CREATININE), and Chloride (CHLORIDE). Note that we had focused on changes in renal function, as suggested by the expert and as was the goal of the HMO to which the patients belonged.

The prediction *task* was to identify patients who, in the second year after their first observation, will have normal Albuminuria (58 patients) versus patients who will manifest micro-albuminuria or even macro-albuminuria, (93 patients). Albuminuria is one indicator of renal failure.

## 5.2. Experimental Design

For each clinical domain, i.e., the Oncology, Hepatitis, and Diabetes domains, the analysis included a series of computational experiments. Each experiment focused on solving a similar classification or prediction task for the domain, using the iDTW similarity measure, in different settings of dependent, adjustable variables, which might have affected the underlying matching process and its results. These variables, or concepts, used in the distance-measure function, were either all raw-data concepts, or all temporal abstractions.

Two KBTA abstraction types were used to evaluate the expected significance of using one or more knowledge-based abstractions in comparison to raw data: (1) the *State* abstraction; and (2) the *Gradient* abstraction. The abstractions were either what we have referred to as *unidimensional* (either a State or a Gradient abstraction of each concept) or what we have referred to as *multidimensional* (both State and Gradient abstractions of the same concept within the same record).

When segmenting the timeline of each record using the GMS method within the GBR process, we used a temporal granularity of $G$ = Day for Hepatitis and Oncology, and a temporal granularity of $G$ = Month for diabetes, due to the more sparse nature of the data in that domain.

To determine the classification of a patient, and directly exploit the iDTW distance measure, we used the *K-Nearest-Neighbors* (*KNN*) technique [Fix and Hodges, 1951]. To evaluate the new unlabeled observation (i.e., patient classification), we used a posterior probability, for KNN classification [Atiya, 2005; Jirina and Jirina, 2011; Silverman, 1986] which defines the probability that a new observation $x$ belongs to class $C_m$ as the fraction ($K_m$) of its k-nearest-neighbors ($K$) that belongs to class $C_m$: $\hat{P}(C_m|x) = \frac{K_m}{K}$. To limit the selection of $k$ neighbors, we used as an upper bound the square root of the number of the training set observations ($\sqrt{n}$ neighbors), an often-used heuristic rule of thumb for the approximation of the optimal *k* [Bichler et al., 2004; Hassanat et al., 2014; Jirina and Jirina, 2011; Mujumdar and Kumar, 2012; Perl and Wohlin, 2004]. To break ties in classification, using a majority of the neighbors, only an odd number of (*k*) neighbors (between 1 and $\sqrt{n}$) was used.

The set of dependent variables, permutated over the experiments included:

(1) *Interpolation method*: The three interpolation methods mentioned in Section 4.2.3 - *Nearest Neighbor, Linear*, and *Average interpolation* (we briefly present in the Discussion section the results of trying out the IBAP interpolation method mentioned in Section 4.2.3);

(2) *Interval-based aggregation method*: The two proposed *duration-based* interval-based aggregation techniques LI, and MTT (see Section 4.2.3), when the duration of two or more intervals within the same time-granule was different. (Note that we always use a *value-based* Delegate Function, for which the default was using the mean value of the samples, in the case of equal sized event intervals (see Section 4.2.3); a special case of this setting was applied to the raw numeric data, which included only instantaneous samples);

(3) *Warping window size*: The two most common temporal warping windows constraints used when applying the DTW similarity measure (10% and Infinity [no constraint]) (we discuss in the Discussion section the results of trying out the KB-Band window-constraining method mentioned in Section 4.3);

(4) *Number of neighbors for classification*: The number of closest neighbors used for classification of an instance (patient) by the KNN classifier, which ranged in our experiments over all odd numbers (to avoid ties when using the majority) from the single closest neighbor (*k*=1), to $k = \sqrt{N}$, where *k* is first rounded off to the closest integer, and *N* being the size of the training set (i.e., in the case of the KNN method, the full data);

(5) *Number of data concepts used as the basis for the matching features*: The number of base medical concepts used for matching (1, 2, or 3 concepts);

(6) *Representation*: The concepts' representation – either raw data only, or a representation of the concepts used for matching that uses some combination of knowledge-based temporal abstractions out of State, Gradient, and the multi-dimensional State and Gradient abstractions for the same concept (4 options for a univariate concept representation [R,S,G, or SG], 10 options for a bivariate concept representation, and 28 for a trivariate concept representation).

To answer our primary research question, we compared, across all iDTW configuration settings, the mean classification performance when using, for the matching function, only raw, numeric, time-stamped data, to the mean classification performance when using unidimensional or multidimensional abstractions, in the case of an input containing either univariate or multivariate time-series. Note that even when using univariate matching, i.e., one base concept, one might still use two *features* for classification, due to the multidimensionality option (i.e., the two different abstractions of the same concept); similarly, for bivariate matching (two base concepts), up to four different features are available, and for trivariate matching (three base concepts), up to six features might be used in the same matching. (Had we generated additional abstractions that are theoretically possible when applying the KBTA method, such as a Rate abstractions, even more features would be available for the same number of input concepts).

Recall that to enable a fare comparison to the standard DTW algorithm, in the case of using raw time-stamped values without any abstraction, the raw numeric concepts values were first standardized to the population's concept values, using a *Z*-score normalization, and then scaled to the [0, 1] interval (Section 4.2.1). Moreover, and as mentioned before, although the raw data included only time-stamped points, we used the iDTW methodology for the preprocessing step of the control set (raw data) as well. Since the raw, point-based data is in fact a specific use case of equal length intervals (e.g., of Temporal Granularity = Minutes), the iDTW Value-Based aggregation was used and specifically the Mean delegate function which is suggested as default delegate function for numeric values (Section 4.2.3.).

This comprehensive analysis produced a large set of experiments, one per each variable configuration, which we then used to evaluate the answer to the primary research question. Specifically, the total number of experiments performed *per domain*, over all configurations, was:

$$\left(\sum_{k=1}^{max\_concepts} \binom{C}{k} * (3^k + 1)\right) * (i * a * w * n)$$

The first part of the expression, defined by the first brackets, aggregates all combinations possible for the predefined medical concepts and data representations in the domain, for an increasing number of concepts and representations, from all possible selections of one (univariate) concept, each selection having four possible representations, and up to the maximal number of relevant domain-specific concepts (*max_concepts*), in this case three (trivariate), which have $(3^k + 1)$ possible representations out of the overall *C* concepts considered as relevant within each domain. (Note that for any number *k* of base concepts, one can create three (S,G, or SG) abstractions for each concept; thus, an overall number of $3^k$ combinations of abstractions; and there is always and additional one combination containing solely raw concepts). For each combination, we permute the different interpolation functions (*i*); the number of possible aggregation functions (*a*); the number of options for DTW window sizes (*w*); and the number of odd nearest neighbor cardinalities for classification (*n*).

Since several of these factors are constant, such as the two interval-aggregation methods, the only varying factors in the above formulation, among the domains analyzed in our study, are the number of medical concepts used for matching in each domain, from which we decided to chose up to three raw concepts, and the maximal cardinality of the nearest neighbors set (containing only odd numbers), used for classification, which was determined by the number $\sqrt{N}$, where *N* is the number of patient records used in each domain. The other dependent variables in the formulation above remain constant across the three domains (see Table 2).

Table 2: Number of experiments performed for each clinical domain.

|  | C | *max_concepts* | i | a | w | n | Total No. of Experiments |
|---|---|---|---|---|---|---|---|
| **Oncology** | 5 | 3 | 3 | 2 | 2 | 7 | 33,600 |
| **Hepatitis** | 5 | 3 | 3 | 2 | 2 | 6 | 28,800 |
| **Diabetes** | 4 | 3 | 3 | 2 | 2 | 6 | 13,536 |

More concretely, in the Oncology domain, five medical concepts and maximal number of 7 odd neighbors (out of 161 patients in total) were used, leading to 33,600 unique experiments. In the Hepatitis domain, 28,800 experiments were performed, as the same number of medical concepts were used as in the Oncology domain (5), but the maximal number of odd neighbors for classification was limited to 6, since only 125 patients were involved in the matching task. In the Diabetes domain, three medical concepts were used, and 151 patients, which limited the number of odd nearest neighbors for classification to 6, leading to 13,536 unique experiments. Overall, a total of **75,936** experiments were performed: 33,600 in the Oncology domain, 28,800 in the Hepatitis domain and 13,536 in the Diabetes domain.

Each experiment involved the performance of a 10-fold Cross validation, in which the total set of the patients' records was partitioned into ten equal-sized subsets: In each iteration, one of the subsets was defined as the test set (i.e., the target set of patient records to be matched), and the training set to which it was matched, using the iDTW distance measure, and the KNN classification method, consisted of the remaining instances. The estimation of the classification or prediction performance for each experiment (i.e., for each instance of a feature subset and a particular other configuration combination setting) was determined by averaging the performance over all 10 test sets. To represent the performance of each test set we used the AUC metric, a normalized value representing the area under the ROC curve, a commonly used metric for model evaluation and comparison in the machine-learning field. To provide a more comprehensive observation, we also recorded the optimal Specificity and Sensitivity values of each ROC curve, based on Youden's Index [Youden, 1950].

The Gold Standard common to all the experiments was set to the actual class of each patient, which of course was unknown to the proposed analysis methods (for example, whether the patient actually had Hepatitis B or C).

Each research question (the primary one, and all of the secondary questions) was evaluated for its performance by comparing the mean AUC scores of the experimental instances relevant to the research question examined, across all common configurations. (For example, comparing the mean performance of using only one base concept, and only its State temporal abstraction, to the use of only one raw-data concept, across all possible setting combinations, out of the five concepts provided in the case of the Oncology domain [see Table 2]).

Finally, to assess the significance of the difference in performance of *any* two specific cases for *all* research questions, such as in the example just given (i.e., one State abstraction, versus one raw-data concept, in the oncology domain), we used, for each specific comparison, a paired t-test to determine, for the two vectors of all AUC values resulting from the use, in each case, of all other relevant configuration permutations, whether the mean difference between the two vectors, across all pairwise comparisons of values, one from each case, is significant.

We shall report the main results, and in particular those that were found to be significantly different ($P<0.05$).

In the Discussion, we shall summarize and discuss the results. We shall also briefly sum up and discuss the results of assessing the value of several other innovations introduced in our method as potential iDTW configuration settings: The IBAP interpolation method, the KB-Band warping window (the details of these methods, which were mentioned previously in Section 4.2.3 and in Section 4.3, appear in the Appendix), and a comparison of the two duration-based delegate functions (i.e., the two interval-based aggregation methods, LI and MTT). These were not our primary or even secondary research questions, but obviously are of interest within the context of the current study.

## 5.3. Results

### 5.3.1. Interval-Based Knowledge-driven Time-Series Matching Analysis Results

As an overview of the results addressing our primary research question across all three domains, Figure 3 presents a comparison of the mean performance of all of the (knowledge-based) temporal-abstraction combinations to the mean performance of the raw-data-only combinations.

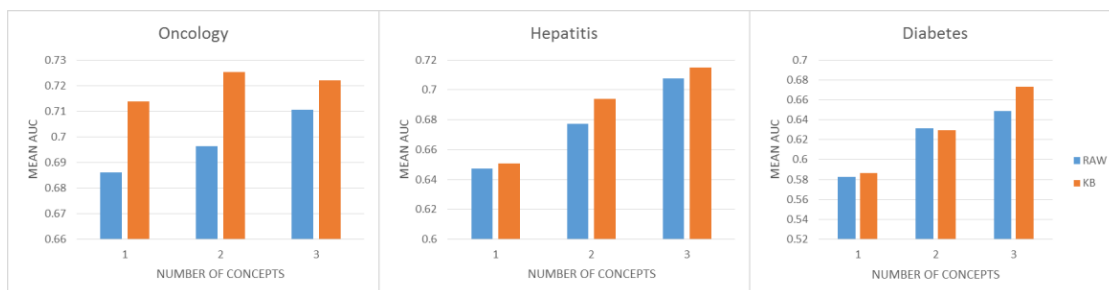

**Figure 3** – The mean classification and prediction performance results of the three datasets, showing the mean AUC when using an increasing number of concepts for matching and classification. KB = knowledge-based temporal abstraction.

The results show a consistent and statistically significant improvement in the mean classification and prediction performance as increasingly more concepts are used for the matching task. (One exception appears in the Oncology domain, when using abstractions, in which the highest performance was reached when using only two concepts, although no significant difference was found between the use of two or three concepts).

**The results shown in Figure 3 demonstrate clearly that, for the purpose of a matching-driven classification, using [only] abstractions is expected, on average, to be at least as good as, and usually better, than using only raw data.**

More concretely, we found that the mean performance of the knowledge-based abstractions was typically significantly superior, compared to the mean performance of the raw data for each dataset, across the three variable-matching options, except in the case of the univariate matching in the Hepatitis domain, and the bivariate matching in the Diabetes domain, in which no significant difference was found between the two representations. Moreover, the variance of the performance scores for all of the knowledge-based abstractions (across all concept subset, representation, and setting configuration combinations) was typically lower, compared to the variance of the raw data for each dataset, across most of the nine major cases (three domains, three variable-matching options within each domain), except in the case of the bivariate and trivariate matchings within the Diabetes domain. For example, in the Hepatitis domain, for the univariate (single) concept selection, across all three abstraction combinations, for all possible configuration settings, the variance in AUC was 0.0078; while when using a single raw concept, the AUC variance, across the same set of possible configuration settings, was 0.0126. Performance variance, in the seven out of nine categories in which abstractions had a lower variance, was usually from 45% to 100% (i.e., twice) higher when using only raw data. Thus, the use of abstractions as features usually seems to lead to more stable results that are less dependent on the particular selection of the base concepts.

### 5.3.2. Multivariate and Multidimensional Classification Analysis Results

To summarize the classification-performance experimental results for each of the three datasets, for each of the three concept-matching cardinality options used (univariate, bivariate, and trivariate), we present in Table 3 an overview of the mean AUC for the different raw or abstract concept combinations used, the combinations that performed the best on average (*Best*), and the combination that performed the worst on average (*Worst*), for each number of concepts.

Recall also that our objective was *not* to achieve or measure any *absolute* performance, but rather, to compare the *relative* merits of the use of different configurations of only the time-oriented data (not even demographic data were included, so as to not obfuscate the results).

The results in all of the domains show a consistently improving (or at least non-decreasing) of the expected performance as increasingly more base concepts were used in the matching phase. This *multivariate* improvement was also found to be statistically significant in all of the cases in which there was an actual increase.

**Table 3** – An overview of the experimental performance results when using various types of abstractions in three different medical domains, using up to three concepts per domain. The performance results of the classification or prediction task for each dataset are shown for each number of variables used within each of the corresponding matching tasks.
S = State, G = Gradient, R = Raw. SG = a multi-dimensional (State and Gradient) abstraction combination for the same variable; "+" indicates a multi-variate combination. The order of the abstractions does not matter.

|  | Univariate | | | Bivariate | | | Trivariate | | |
|---|---|---|---|---|---|---|---|---|---|
|  | Mean AUC | Best on average | Worst on average | Mean AUC | Best on average | Worst on average | Mean AUC | Best on average | Worst on average |
| **Oncology** | 0.71 | SG | R | 0.72 | G+G, G+S, | R+R | 0.72 | S+S+G | SG+SG+G |
| **Hepatitis** | 0.65 | SG | S | 0.69 | SG+SG | S+G | 0.71 | SG+SG+SG | S+G+G |
| **Diabetes** | 0.59 | S | G | 0.63 | S+S | G+G | 0.67 | SG+SG+SG | G+G+G |

Using the *multidimensional* abstraction combination (SG) led to the best performance in most of the matching tasks; it was a component of the worst performance in only one case.

Table 3 also adds another view that answers our main research question, albeit in a different fashion. As can be observed, the best (on average) combinations, are consistently, throughout three domains and three cardinalities of concepts, the abstract-concept combinations (although the only raw-data combinations were not necessarily, on average, the worst). Moreover, it seems that usually, although not always, using State abstractions in these three particular domains led to a better mean performance than using the Gradient abstractions. The ranking of the abstract combinations was found as mostly consistent with this observation throughout the matching tasks of each domain. That is, the best (on average) combinations in the diabetes domain always include the maximal number of State abstractions, and the worst (on average) combinations always consist of Gradients; the best (on average) combinations in the Hepatitis domain include a pure multidimensional combination, and the best (on average) combinations in the Oncology domain include at least one State abstraction.

To further demonstrate the expected classification-performance experimental results for each of the three datasets and answer our primary research question, we present in Table 4 a detailed view of the mean AUC for the different raw or abstract concept combinations used, for the most general case of multivariate matching task in this study – the Trivariate base-concepts matching.

**Table 4** – The mean AUC of trivariate base-concept matching for the 11 possible feature representations (one raw data and nine different abstraction combinations, not considering their order) in three different medical domains, using up to three concepts per domain.

|  | G+G+G | S+S+S | R+R+R | SG+SG+SG | SG+SG+S | S+S+G | SG+S+S | SG+G+G | SG+SG+G | SG+G+S | S+G+G |
|---|---|---|---|---|---|---|---|---|---|---|---|
| **Oncology** | 0.713 | 0.726 | 0.71 | 0.709 | 0.721 | 0.738 | 0.73 | 0.707 | 0.706 | 0.724 | 0.731 |
| **Hepatitis** | 0.702 | 0.718 | 0.708 | 0.736 | 0.733 | 0.699 | 0.728 | 0.712 | 0.722 | 0.715 | 0.692 |
| **Diabetes** | 0.495 | 0.747 | 0.649 | 0.752 | 0.748 | 0.677 | 0.747 | 0.596 | 0.682 | 0.678 | 0.59 |

Table 4 provides yet another perspective to answer our main research question. As can be observed, *on average*, most of the combinations that include abstract concepts performed better, in comparison to the raw data combinations (as it happens, this holds, in each medical domain, for seven out of ten abstract combinations); the differences were also statistically significant. For the nine cases in which the raw combinations performed better than the combinations that include abstract concepts, the difference was also statistically significant, except for two out of three abstract combinations in the Oncology domain. However, as mentioned above, the internal ranking of the different abstractions within each domain varied for each domain.

### 5.3.3 Examples of Instances of the Experimental Results

Although the main research question focuses on the expected performance of interval-based temporal abstractions of the data of several classification and prediction tasks, compared to their expected performance when using raw-data only, we also present in Table 5, for better clarity, and for better understanding of the flavor of our experimental results, the actual settings of the absolute (as opposed to mean) top performing experiments in three different medical domains, for all concept-matching cardinality options (univariate, bivariate, and trivariate).

Recall that each case of the primary or secondary research questions (e.g., the mean performance when using, for matching records, one base concept and only a *State abstraction* in the Oncology domain, in comparison to matching one *raw-data* concept in the Oncology domain) was tested using the mean AUC scores resulting from the use of, say, one State abstraction, across all of the possible configuration settings.

However, as opposed to our usual mean-case analysis, here we shall examine, for the first time, several instances of the actual absolute performance of specific concept-subset and configuration-setting combinations.

In Table 5, we listed the top univariate, top bivariate, and top trivariate absolute performing instances (whether raw or abstract) in each domain, and their relative ranking, across all combinations of concept subsets and configuration combinations, within that domain. (Each instance includes a specific set of configurations that is not shown.)

Note that the number of raw-data experiments was in a sense doubled artificially, and thus, respectively, the number of corresponding resulting instances. The reason is that in the case of the two *duration*-based aggregation function options, LI and MTT, which are a part of the configuration-settings options, the result of aggregating several raw data points, whose duration is always equal, is the same regardless of the configuration setting. That is due to the fact that when the durations are equal, we use the *value*-based aggregation function (whose default was the mean value) (see Section 4.2.3). We kept these configurations for the sake of simplicity and completeness, so the experimental setup was identical for raw and for abstract concepts. (Of course, identical instances might sometimes result also from applying the two different aggregation functions to abstract concepts, especially when they are also point based, or include only one interval within each time-granule). Thus, the number of *unique* raw data instances in the top performance ranks is actually half of what we are listing. We shall ignore this small detail when describing the examples, even though taking it into consideration might further strengthen our results.

As can be observed in Table 5, in two out of three medical domains (Hepatitis and Diabetes), the top performing experiments (combinations) are abstract-concept combinations, both of which happen to bivariate, while in the third domain (Oncology), the top combination was a raw trivariate one (although the mean performance of abstract-concept combinations was higher than the mean performance of raw-concept combinations). In the Diabetes domain, the highest score was obtained by an abstract-concept combination whose mean AUC was 0.894 (see Table 5). (In this domain, when looking at the complete results, the best raw-data combination mean AUC was 0.882, and was ranked in the 58[th] place.) In the Hepatitis domain, the highest score was obtained by an abstract-concept combination whose mean AUC was 0.88 (see Table 5). (In this domain, when looking at the complete results, the best raw-data combination mean AUC was 0.858 and was ranked in the 33[rd] place). In the Oncology domain, the highest score was obtained by a raw-data combination whose mean AUC was 0.931 (see Table 5). (In this domain, when looking at the complete results, the best abstract-data combination mean AUC was 0.923 and was ranked in the 29th place).

**Table 5** – A snapshot from the individual experimental instances' results in three different medical domains, showing the top performing instance for each base-concept cardinality (univariate, bivariate, and trivariate), using up to three concepts per domain, and their relative rank [with respect to the AUC] within their domain. For clarity, the value of the instance's configuration settings, such as the Interpolation method used, are not presented.
S = State, G = Gradient, R = Raw. SG = a multi-dimensional (State and Gradient) abstraction combination for the same variable; "+" indicates a multi-variate combination. See Section 5.1 for the concept-name abbreviations.

| Oncology | | | | Hepatitis | | | | Diabetes | | | |
|---|---|---|---|---|---|---|---|---|---|---|---|
| Concepts | Representation | Mean AUC | Rank | Concept | Representation | Mean AUC | Rank | Concept | Representation | Mean AUC | Rank |
| WBC+ HGB+ BANDS | R+R+R | 0.931 | 1 | LDH+ ALP | SG+S | 0.88 | 1 | ALBUMINURIA-U24h + CHLORIDE | SG+SG | 0.894 | 1 |
| WBC+ PLATELET | R+R | 0.929 | 3 | LDH+ ALP+ I-BIL | S+S+G | 0.864 | 7 | ALBUMINURIA-U24h + CREATININE+ CHLORIDE | SG+S+S | 0.893 | 3 |
| WBC | R | 0.925 | 9 | LDH | R | 0.853 | 51 | ALBUMINURIA-U24h | SG | 0.885 | 27 |

When considering the full results in the Diabetes and Hepatitis domains, the overall dominance of the instances including abstract concepts, and specifically those including multidimensional abstract-concepts, was evident, even beyond the combinations that were ranked as first, which appear in Table 5.

For example, in the Diabetes domain, when considering separately the univariate, bivariate, and trivariate cases, the top-performing 4, 54, and 3,323 combinations, respectively, were all abstract-concept combinations. Out of these, 100%, 72%, and 77% of the combinations, respectively, included at least one multidimensional abstract-concept.

In the Hepatitis domain, when considering separately the bivariate and trivariate cases, the top-performing 16 and 18 combinations were all abstract-concept combinations. Out of these, 87% and 55% of the combinations, respectively, included at least one multidimensional abstract-concept. The univariate matching task's top four combinations were raw ones.

In the Oncology domain, when considering separately the univariate, bivariate, and trivariate cases, the top-performing 6, 10, and 16 instances in the Oncology domain were raw concepts.

Typically, in each domain there were concepts that were consistently featured among the best performing combinations, throughout all of the matching tasks. In the case of the Diabetes domain, that was the ALBUMINURIA-U24h concept; in the case of the Oncology domain, it was the WBC concept; in the case of the Hepatitis domain, these were the LDH and ALP concepts, followed by I-BIL, as can also be noticed in Table 5.

Note again that using the knowledge-based temporal abstractions can usually outperform *on average* the use of only raw data, although in specific case, the top-performing combinations might include raw-data concepts; and that the specific abstractions or raw-data concepts might need to be carefully chosen within each domain.

### 5.3.4 Mean Optimal Sensitivity and Specificity Analysis Results when Using Different Representation Methods

A more detailed analysis of the performance of the different combinations, considering the trade-off between the sensitivity and specificity measures, produced results that were similar to the average AUC performance results, although a definite tradeoff between the two measures appeared for several combinations. For instance, in all three medical domains when applying the multivariate matching task (i.e., bivariate and trivariate concept matching), the *sensitivity* achieved by the use of abstractions was consistently *higher* than that achieved by the use of only raw data, while for both the Oncology and the Hepatitis domains, the mean optimal *specificity* of the use of abstractions was consistently *lower* than the specificity of using only raw data (which was ranked in the middle among all abstraction combinations, in the case of the diabetes domain). Figure 4 presents detailed results in the Oncology domain for trivariate matching. (Interestingly, using, in this particular case, only three abstract features, generated from the three base concepts, resulted on average in a better performance than when using a higher number of generated abstractions – four, five and even six [when using two abstraction types for each of the three concepts.)

These results might be explained by the fact that the temporal abstraction process discretizes the data by mapping a range of values into a single value, creating a smaller set of values. Thus, two records might seem more similar than they actually are, increasing the sensitivity, but also potentially decreasing the specificity.

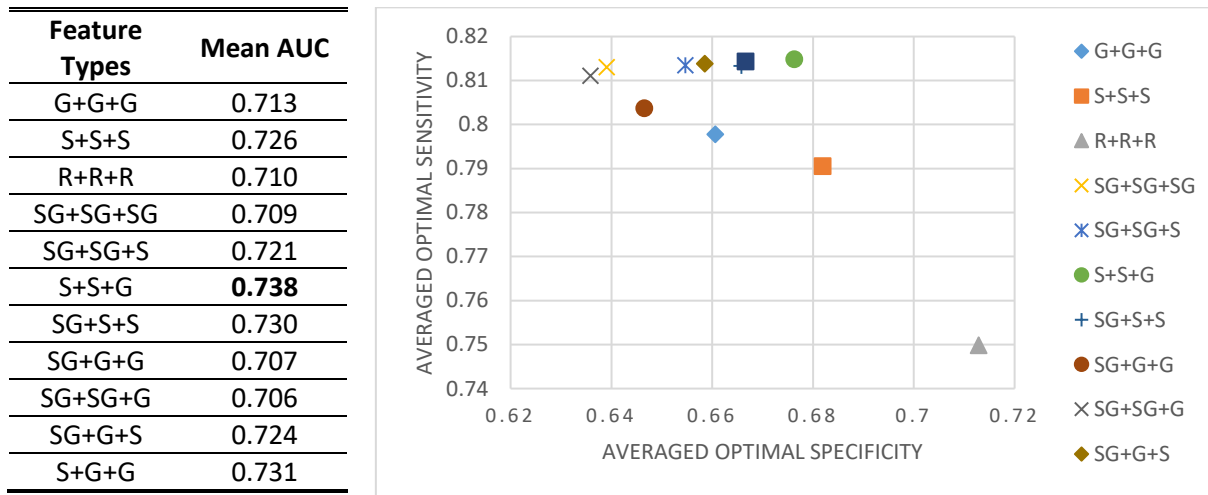

| Feature Types | Mean AUC |
|---|---|
| G+G+G | 0.713 |
| S+S+S | 0.726 |
| R+R+R | 0.710 |
| SG+SG+SG | 0.709 |
| SG+SG+S | 0.721 |
| S+S+G | **0.738** |
| SG+S+S | 0.730 |
| SG+G+G | 0.707 |
| SG+SG+G | 0.706 |
| SG+G+S | 0.724 |
| S+G+G | 0.731 |

**Figure 4** - Mean optimal Sensitivity and Specificity of the ROC curves of different abstraction types, produced by each experimental instance, for all of the combinations of settings for the trivariate concept matching, in the Oncology domain, and the mean AUC of trivariate concept matching for raw data and for all abstraction types in the Oncology domain, averaged over all of the configurations.
S = State, G = Gradient, R = Raw. SG = a multi-dimensional (State and Gradient) abstraction combination for the same variable; "+" indicates a multi-variate combination.

### 5.3.5 The Effect of the Neighborhood Size on the KNN Classification and Prediction Performance

Recall that as part of our experimental procedure (See Section 5.2), we had assessed the iDTW mean classification performance over increasing number of *k*-nearest neighbors, comparing the record to be classified to its single closest neighbor (*k*=1), and up to its $k = \sqrt{N}$ closest neighbors, including only the odd numbers, to break ties. Thus, we were able to analyze the mean performance when using abstract combinations for matching compared to when using only raw data combinations, as an increasing number of neighbors were used for the actual classification.

As expected, across all configurations, the use of an increasing number of neighbors monotonically increased the classification and prediction performance, until reaching an optimal saddle-point around the square root of the training set size *N*, regardless of the clinical domain being used or the matching problem applied.

However, in quite a few cases in which no significant improvement was found when an additional number of neighbors were used, across all configurations, it seemed that the reason was due to the fact that performance when using abstraction-based concepts reached, in about half of the cases, an early plateau, as opposed to the consistent improvement of the performance when using only raw data. *In fact, in those cases, multiple abstract combinations reached a mean optimal performance plateau very early, using only as few as five neighbors,* while the performance using only the raw data continuously improved as an increasingly larger number of neighbors was used, up to the maximal possible number of neighbors.

Moreover, *we could not find any single matching task in which using only the raw data reached a saddle point earlier than when using the best (on average) abstract concept combination*.

Figure 5 presents a typical instance of the phenomenon we are describing: the KNN classification results for the univariate concept matching configurations in the Diabetes domain. While the performance when using only raw data continues to improve as a larger number of nearest neighbors is used, the performance of the best abstractions-based matching reaches its maximum very early, when $k = 5$.

We discuss the possible meaning of this result in the Discussion section.

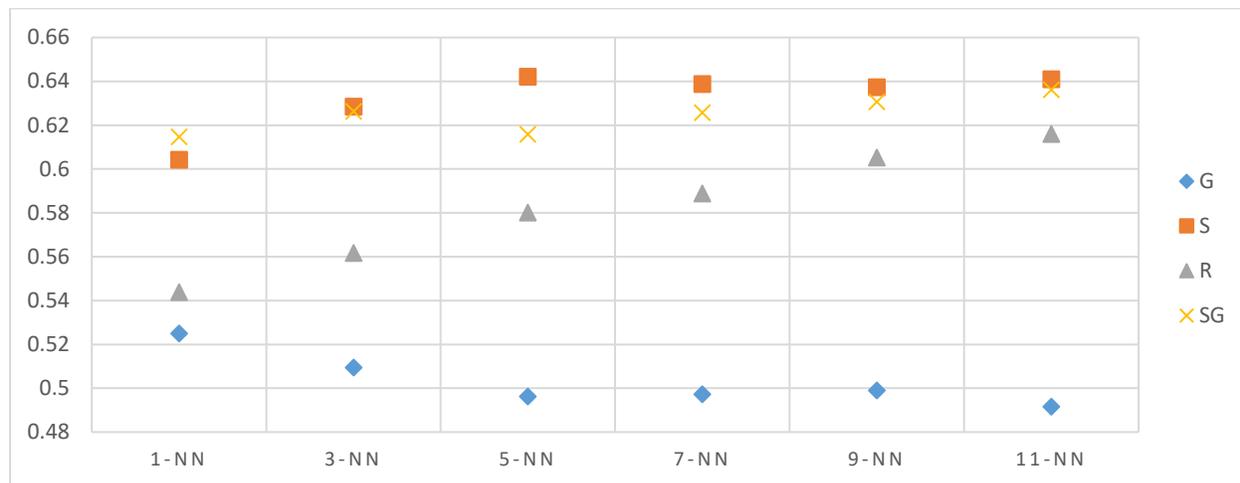

**Figure 5** - The *k* Nearest Neighbors (KNN) classification results for the **univariate** matching problem when using different abstraction types in the **Diabetes** domain. The Y-axis represents the mean AUC; the X-axis represents the number of nearest neighbors used. S = State, G = Gradient, R = Raw. SG = a multi-dimensional (State and Gradient) abstraction combination for the same variable; "+" indicates a multi-variate combination.

## 6. Summary and Discussion

Finally, we shall now discuss in detail both the contributions of this study, as well as its limitations.

### 6.1 Key Methodological Contributions and Results

We designed, implemented, and evaluated a new generic time-series matching methodology, iDTW, for determining the similarity between two or more multivariate interval-based longitudinal electronic health records that enhances the standard DTW time-series matching algorithm, to support both a multivariate and an interval-based representation of the data, and thus defined a similarity measure that exploits the interval-based, knowledge-driven abstractions of the raw data. *Our experiments have demonstrated a significant improvement on average and a consistent superiority of the expected performance (averaged over all variable subsets and methodological configuration settings) when* *using one or more knowledge-based temporal abstractions, compared to the expected performance when using only raw data, with respect to the same matching-based classification and prediction tasks.*

*Our experiments have also shown that a lower variance in performance was typically observed when using the knowledge-based abstractions for matching, compared to the variance in performance when using only raw data.* This conclusion, together with the former one, indicate that using abstract concepts is expected, *on average*, to both perform better for matching tasks, and in a more stable manner.

Specifically, *in each medical domain, there existed at least one abstraction, or a combination of abstractions, whose use during the matching process led, on average, over all concept subsets and iDTW methodological configurations, to a significantly better performance than the use of only raw data.* However, the precise abstraction combination and configuration setting might still need to be determined for each domain. The particular set of experiments we performed demonstrated the superiority of the use of abstract concepts: In two out of three medical domains (Hepatitis and Diabetes), all of the [multiple] top performing feature [concept] combinations were consistently abstract-concept combinations. In particular, the use of pure multidimensional combinations (i.e., using the SG combination for each of the concepts used) almost always outperformed the use of only raw data, regardless of the medical domain used or the number of variables used in the matching task (univariate, bivariate, or trivariate). The only exception was the trivariate matching task in the Oncology domain, in which the use of pure multidimensional abstractions was found, on average, to be slightly inferior to the use of the raw data, although no significant difference was found. The dominance of the multidimensional combinations was also noted when analyzing the individual experiments for each domain, where most of the top abstract combinations included at least one multidimensional abstract-concepts.

Note that our precise result is that for all domains, and classification or prediction tasks, and variable numbers (univariate, bivariate, or trivariate), the *mean* performance, averaged over all concept (variable) subsets, and over all method-configuration (e.g., interpolation method, window duration) combinations, is always higher when using abstractions, than when using only raw data. However, *this result does not reveal which specific concept subset and method configuration might be best for each domain (or even domain and task)*. In particular, for each task (e.g., prediction of survival, versus detection of an underlying complication) or domain (e.g., Oncology versus ICU), different combinations of features are likely to provide the best classification and prediction performance. It might even occur (as was shown in the Results for the domain of oncology, for the task of determining which type of bone-marrow transplantation was performed) that the top-performing data set might happen to be composed of only raw data (even though the *average* performance of abstract concepts, over all configurations, was still superior).

Thus, one way of looking at our results is that if, *given a new classification or prediction task, one can only try out a limited number of feature combinations and configuration-setup options, it might be better to use temporal-abstraction features*. On average, they will lead to a *higher mean performance*, accompanied by a *lower variance* between the performance of different base-concept and configuration-setup choices, thus *reducing the risk* of missing the optimal option by a large margin. Using temporal abstractions might also reduce the search space when performing a satisficing random search, if the search focuses on, or is at least biased towards abstractions.

Furthermore, the nature of the data and the constraints imposed by the KBTA methodology might play a part as well. For example, when adjacent time points are too far apart (as might be the case for "slow" domains with relatively slow sampling rates, such as Diabetes or Hepatitis) no Gradient abstractions might be formed with any certainty. But features including State abstractions and their long-duration interval-based interpolations might well provide solid insights. On the other hand, when adjacent values of the same variable are too close, gradient values will be formed, but might vary too often (a situation in which we can use other abstraction types, such as Trends, which we had shown to be quite useful in the case of the ICU domain and the task of predicting Sepsis [Sheetrit et al., KDD-2019]). Alternatively, in the case of continuously sampled data, such as hematological data in the first month or two following a bone-marrow transplantation event, without a lot of missing data (as was the case in this study for the oncology domain data set), raw-data subsets might well perform well, due to a reduced need for abstraction and interpolation.

The benefit of using knowledge-driven abstractions became even clearer in several cases in which, when using abstractions for matching, fewer training samples (nearest neighbors) were needed, on average to reach, the mean optimal KNN-based classification performance. In fact, as the example shown in Figure 5 has demonstrated, sometimes only five closest neighbors were sufficient to reach the performance plateau when using abstractions, while when using raw-data concepts, performance continued to improve until the maximal possible number of $k = \sqrt{N}$. On the other hand, when using only raw data, a saddle point was never reached earlier than when using the best (on average) abstract concept combination. One way of looking at this result is that it again demonstrates the reduction in variance among different configuration choices, when using temporal abstractions, the configuration choice in this case being the value of $k$. This phenomenon further strengthens our study's main original hypothesis, namely that using knowledge-driven temporal abstractions might enhance the matching process: In this case, it might suggest that when using abstract concepts, the noise and the number of outliers in the data are reduced, and thus, often a smaller number of neighbors might be sufficient to correctly classify a test sample.

In a way, our current study can be seen as extending the Keogh and Pazzani [2001] derivative-based approach, which matches time series using only local derivative values (which represent an abstraction of the *shape* of two wave forms) into general temporal abstractions, such as State abstractions, in addition to local Gradient abstractions. Thus, we are using not just an abstraction of the *direction*, but also an abstraction of the actual *values*; and we are adding also longer, interval-based *durations* (and not just point-based local derivatives).

Analyzing the expected classification and predication performance within the three clinical domains shows that using an increasing number of concepts (i.e., three instead of two, and two instead of one) usually, in eight out of the nine categories (i.e., domain and number of base concepts) of experiments that we have conducted (see Table 3), increases the mean accuracy, regardless of the domain or concept types in use (and in no category did it significantly decrease performance). This phenomena was also reflected when further analyzing the experimental results, in which the top performing experiments included multivariate concepts (see Table 4), although in two out of three domains (Hepatitis and Diabetes) the best experimental result actually consisted of only two concepts (bivariate) instead of the maximal number of three concepts (trivariate). Thus, although the original, classic DTW algorithm was restricted to the univariate matching problem, its generalization to the multivariate matching problem, through the granularity-based temporal-segmentation procedure we had introduced, proved to be feasible and efficient, and suitable for our multidimensional task.

However, one limitation of this study is that we have limited the maximal number of features to *six* – three clinical concepts, each potentially abstracted into two abstractions (State and Gradient). Although the results were encouraging, it would be interesting and useful to examine in an even more complex and higher dimensional environment.

Although abstractions were in general found superior, on average, in most cases, different abstraction types performed differently, depending on the domain in which they were used. For instance, in the Diabetes domain, regardless of the number of concepts used in the matching task (univariate, bivariate, or trivariate), the use of State abstractions dominated, on average, the use of all other abstractions or the use of raw data. However, the use of State abstractions was found inferior, on average, to the use of some of the other abstraction combinations in both the univariate and bivariate matching tasks for the two other clinical domains, in which, perhaps, the *trend* is at least as important as the amplitude of the value.

Similar conclusions were inferred from the more detailed analysis of the performance of the different combinations, when examining closely the trade-off (on average) between the *sensitivity* and the *specificity* measures. Typically, the mean optimal *sensitivity* achieved when using abstractions for matching was consistently *higher* than the mean optimal sensitivity achieved when using only raw data, while the mean optimal *specificity* when using abstractions was consistently *lower* than the mean optimal specificity when using only raw data. These results might be explained by the fact that the temporal abstraction process discretizes the raw data values and trends by mapping, in each case, a range of values into a single value (e.g., LOW or INC), creating a smaller set of values. This value restriction would tend to make two records with a similar underlying classification seem more similar and thus would increase *sensitivity*, but would also decrease *specificity*.

The fact that, on one hand, specific knowledge-based temporal-abstraction combinations performed, on average, better than when using only raw data within each domain, and on the other hand, in some cases a tradeoff between the sensitivity and specificity exists, suggests future studies that are aimed at determining the best combination of temporal abstractions to be used in each domain and task, through extensive experimentation of the type we had performed. Furthermore, the optimal combination might be chosen, depending on the *preferences* (*utility function*) of the user with respect to being more sensitive or more specific.

Another future direction to explore is the use of the Pattern abstraction, which in the KBTA ontology indicates a temporal relationship between different intervals composed of different types. Using temporal patterns – either predefined or discovered in a data-driven fashion - could diminish dramatically both the dimensionality of the matching task and its running time. We have successfully used the existence, cardinality within each longitudinal record, or mean duration, of frequent temporal patterns, as classification features [Moskovitch and Shahar, 2015b], as well as their overall distribution [Sheetrit et al., 2019]. However, it might also be beneficial to match the temporal patterns of a pair of longitudinal records in a time-oriented fashion, as is done in the iDTW methodology.

## 6.2 Additional Innovative Aspects of the iDTW Methodology

Although they were not a part of our primary or even secondary research questions, we did examine the value of several other innovations introduced in our method as iDTW parameters: The *Interval-Based Adjacent Proportional interpolation* (IBAP) interpolation method, the *knowledge-based Band* (KB-Band) warping window (see Appendix for details regarding these methods), and (as part of the main experiments) the two interval-based aggregation methods that were a part of the standard iDTW configuration.

In particular, we had examined the expected performance (again, averaged over all concept and configuration combinations) of the IBAP interpolation method that we had proposed here. In general, the various interpolation functions, including the IBAP method, performed differently in each domain, and also in the different matching tasks. Thus, it seems that the optimal interpolation technique to be used depends on the domain. Additional details can be found elsewhere [Lion, 2015].

As to restricting the matching process during the iMatch phase, in general, the classification and prediction performance of both the restricted 10% window and of the KB-Band variations (see Section 4.3) usually outperformed, on average, the unrestricted variation in both runtime and average performance. However, the ranking of the two restricted variations was inconsistent and depended on the domain and in some cases also on the specific matching task applied. For example, In the Hepatitis domain, the KB band was found superior in both of the multivariate matching tasks. These results support Ratanamahatana and Keogh's [2004] claim that a narrow constraint is preferable when matching time-series. However, while Ratanamahatana and Keogh evaluated only univariate, regularly sampled, point-based time series, our study extends the evaluation to the use of multivariate, irregularly sampled, interval-based time-series.

Note that the KB-Band is defined by the actual properties of the data, while the 10% and the unrestricted matching variation are methods used by default in many studies. The fact that it can outperform the standard variation using knowledgeable reasoning is encouraging. Furthermore, the KB-Band notion can open a new window of opportunities for future studies, in a variety of domains, not only in the medical domain, in cases in which the preferable window size is unknown, and thus save heavy experimental efforts in order to optimally set its size.

Finally, we had also examined, as part of performing the main experiments, the expected performance (averaged over all concept and configuration combinations) of the two interval-based aggregation techniques that we used for representing a time period by a single value, also referred to as a delegate value. (See Section 4.2.3): (1) The *Maximal Total Time* (MTT) function -- the maximal cumulative duration of all of the applicable intervals during the relevant aggregation time period, and (2) The *Longest Interval* (LI) -- the interval that has the longest duration during that time period. In the case in which all intervals were of the same length, the Average of the values of the concepts holding over the intervals was used. No significant difference was found between the two interval-based aggregation techniques in two out of the three datasets used (Oncology and Hepatitis), regardless of the number of concepts used, although the MTT method was found significantly superior in the Diabetes dataset when one or two concepts were used for the matching task, and the LI method prevailed when three concepts were used. The importance of using one or the other of the interval-based aggregation methods in that domain might be due to the sparse nature of the diabetes database, which led to the use of a temporal granularity of months (versus days, in the other domains), and to a larger number of intervals per time-granule. It might well be that using a different, coarser temporal granularity $G$ for the other two domains (e.g., Month instead of Day) might have produced differences between the use of LI and MTT as well.

As in the case of the choice of the interpolation function, it seems that the performance of these methods depends on the specific domain in which they are used, and they should be used after carefully considering each domain's characteristics.

Furthermore, as noted in Section 5.3.3, choosing the LI or MTT functions often made no difference, such as in the case of raw data, or of point-based abstractions, or of single interval-based abstractions, etc., thus possibly reducing the statistical difference between them, which might in reality exist if only relevant instances, in which the choice makes a difference, are considered.

The evaluation performed in this study included only three relatively small datasets from three different clinical domains, Oncology, Infectious Hepatitis, and Diabetes, and thus suffers from all of the limitations of such limited empirical studies. Naturally, there are many other medical domains in which our results could be validated. Furthermore, other iDTW parameters included in our evaluation, whose value was not examined in detail in the current study (e.g., the warping window size, or the use of the IBAP methodology) could affect the classification and prediction performance.

It is also worth noting that, although the temporal-abstraction knowledge itself, for each clinical concept, was provided by medical experts, involving in the future domain experts in assessing the meaning of the combinations that eventually proved useful in our more than 70,000 experiments, could shed even more light on our findings and can direct future studies.

## 7. Conclusions

The use of interval-based knowledge-based temporal abstractions significantly enhanced the mean performance of time-series matching tasks, and decreased the variance in performance among different concept-subset and configuration-setup choices, although optimizing performance requires choosing the abstractions carefully. Using multi-dimensional abstractions, i.e., exploiting several different abstractions of the same concept when performing the matching task, usually improved the accuracy of the classification or predication tasks, in comparison to the use of either only raw, time-stamped aggregated data or only uni-dimensional abstractions. The use of abstractions also seems to considerably reduce the mean number of nearest neighbors needed for classification by a KNN procedure.

The current study has also demonstrated a successful generalization of the classic Dynamic Time Warping matching algorithm for the multivariate matching task, especially when using abstract features; increasing the number of features used improved the mean performance of almost all classification and prediction tasks.

**Acknowledgments**

We would like to acknowledge all of the clinicians who had contributed from their time and knowledge in assisting us in acquiring the three domain knowledge-bases used for this evaluation work.

**Funding:**

This work was partially supported by the Israeli National Institute of Health Policy Research, Grant No. 2015/111/A and ONR award No. N629091912124.

# REFERENCES

Aach, J,. and Church, G. M. (2001). Aligning gene expression time series with time warping algorithms. *Bioinformatics*, *17*(6), 495–508.

Azulay, R., Moskovitch, R., Stopel, D., Verduijn, M., de Jonge, E., and Shahar, Y. (2007). Temporal Discretization of medical time series - A comparative study. In *Proceedings of the 11th International Workshop on Intelligent Data Analysis in Medicine and Pharmacology IDAMAP*.

Berndt, D. and Clifford, J. (1994). Using dynamic time warping to find patterns in time series. *AAAI-94 Workshop on Knowledge Discovery in Databases (KDD-94)*, Seattle, Washington.

Brubaker, J. R., Lustig, C., and Hayes, G. R. (2010). PatientsLikeMe: empowerment and representation in a patient-centered social network. In *CSCW'10; Workshop on Research in Healthcare: Past, Present, and Future.*

Chaovalitwongse, W. A., Fan, Y. J., and Sachdeo, R. C. (2007). On the time series k-nearest neighbor classification of abnormal brain activity. *Systems, Man and Cybernetics, Part A: Systems and Humans, IEEE Transactions*, *37*(6), 1005–1016.

Chen, L., and Ng, R. (2004). On The Marriage of Lp-norms and Edit Distance. In *Proceedings of the Thirtieth International Conference on Very Large Data Bases* (Vol. 30, pp. 792-803).

Chen, L., Özsu, M. T., and Oria, V. (2005). Robust and Fast Similarity Search for Moving Object Trajectories. In *Proceedings of the 2005 ACM SIGMOD International Conference on Management of Data* (pp. 491-502). ACM.

Combi, C., and Shahar, Y. (1997). Temporal reasoning and temporal data maintenance in medicine: Issues and challenges. *Computers in Biology and Medicine* 27 (5) 353–368.

Ewing, R. M., Kahla, A. B., Poirot, O., Lopez, F., Audic, S., and Claverie, J. M. (1999). Large-scale statistical analyses of rice ESTs reveal correlated patterns of gene expression. *Genome Research*, *9*(10), 950–959.

Faloutsos, C., Ranganathan, M., and Manolopoulos, Y. (1994). Fast subsequence matching in time-series databases Matching in Time-Series Databases. *SIGMOD '94 Proceedings of the 1994 ACM SIGMOD International Conference on Management of Data* (Vol. 23, No. 2, pp. 419-429). ACM.

Forestier, G., Petitjean, F., Riffaud, L., and Jannin, P. (2015). Optimal Sub-Sequence Matching for the Automatic Prediction of Surgical Tasks. *Artificial Intelligence in Medicine* (pp. 123-132). Springer International Publishing.

Giorgino, T., Tormene, P., and Quaglini, S. (2007). A multivariate time-warping based classifier for gesture recognition with wearable strain sensors. In *Engineering in Medicine and Biology Society, 2007. EMBS 2007. 29th Annual International Conference of the IEEE* (pp. 4903-4906). IEEE.

Goldstein, A., and Shahar, Y. (2016). An automated knowledge-based textual summarization system for longitudinal, multivariate clinical data. *The Journal of Biomedical Informatics* 61:159-175.

Goldstein, A., Shahar, Y., Orenbuch, E., and Cohen, M. (2017). Evaluation of an automated knowledge-based textual summarization system for longitudinal, multivariate clinical data. *Artificial Intelligence in Medicine* 82:20–33.

Häne, B. G., Jäger, K., and Drexler, H. G. (1993). The Pearson product-moment correlation coefficient is better suited for identification of DNA fingerprint profiles than band matching algorithms. *Electrophoresis*, *14*(1), 967-972.

Ho, T. B., Nguyen, T. D., Kawasaki, S., Le, S. Q., Nguyen, D. D., Yokoi, H., and Takabayashi, K. (2003). Mining hepatitis data with temporal abstraction. In *Proceedings of the Ninth ACM SIGKDD International Conference on Knowledge Discovery and Data Mining* (pp. 369-377). ACM.

Hripcsak, G, Albers, D. Perotte, A. (2015). Parameterizing time in electronic health record studies. *Journal of the American Medical Informatics Association* 22 (4), 794-804.

Keogh, EJ and Pazzani, MJ (2001). Derivative Dynamic Time Warping. Proceedings of the *2001 SIAM International Conference on Data Mining* (SDM).

Keogh, E. and Ratanamahatana, C. A. (2005). Exact indexing of dynamic time warping. *Knowledge and Information Systems*, *7*(3), 358–386.

Keogh, E., Xi, X., and Wei, L. (2006). The UCR Time Series Classification/Clustering - http://www.cs.ucr.edu/~eamonn/time_series_data.

Klimov, D., Shahar, Y., and Taieb-Maimon M. (2009). Intelligent visualization of temporal associations for multiple time-oriented patient records. *Methods of Information in Medicine* 48(3), 254-262.

Klimov, D., Shahar, Y., and Taieb-Maimon M. (2010a). Intelligent querying, visualization, and exploration of the time-oriented data of multiple patients. *Artificial Intelligence in Medicine* 49, 11-31.

Klimov, D., Shahar, Y., and Taieb-Maimon M. (2010b). Intelligent selection and retrieval of multiple time-oriented records. *The Journal of Intelligent Information Systems* 35, 261-300

Kostakis, O., Papapetrou, P., and Hollmén, J. (2011). ARTEMIS: Assessing the similarity of event-interval sequences. *Machine Learning and Knowledge Discovery in Databases* (pp. 229-244). Springer Berlin Heidelberg.

Kotsifakos, A., Papapetrou, P., and Athitsos, V. (2013). IBSM: Interval-Based Sequence Matching. In *SDM* (pp. 596-604).

Kovács-Vajna, Z. M. (2000). A fingerprint verification system based on triangular matching and dynamic time warping. *IEEE Transactions on Pattern Analysis and Machine Intelligence*, *22*(11), 1266–1276.

Lin, J., Keogh, E., Wei, L., and Lonardi, S. (2007). Experiencing SAX: a novel symbolic representation of time series. *Data Mining and Knowledge Discovery*, *15*(2), 107–144.

Lion, M. (2015). Interval-based, knowledge-driven similarity measures for the retrieval of longitudinal medical records. Thesis submitted in partial fulfillment of the requirements for the Master of Sciences degree, Department of Information Systems Engineering, the faculty of Engineering Sciences, Ben Gurion University of the Negev, Israel.

Mörchen, F., and Ultsch, A. (2005). Optimizing time series discretization for knowledge discovery. In *Proceeding of the Eleventh ACM SIGKDD International Conference on Knowledge Discovery in Data Mining* (pp. 660-665). ACM.

Martins S.B., **Shahar Y.**, Goren-Bar D., Galperin M., Kaizer H., Basso L.V., McNaughton, D., and Goldstein, M.K. (2008). Evaluation of an architecture for intelligent query and exploration of time-oriented clinical data. *Artificial Intelligence in Medicine* **43**, 17-34.

Moskovitch, R. and Shahar, Y. (2015a). Fast time intervals mining. *Knowledge and Information Systems* 42:21-48.

Moskovitch, R. and Shahar, Y. (2015b). Classification of multivariate time series via temporal abstraction and time intervals mining. *Knowledge and Information Systems* 45 (1):35-74.

Moskovitch, R. and Shahar, Y. (2015c). Classification-driven temporal discretization of multivariate time series. *Data Mining and Knowledge Discovery* 29 (4):871-913.

Park, S., and Kim, S. W. (2006). Prefix-querying with an L1 distance metric for time-series subsequence matching under time warping. *Journal of Information Science*, *32*(5), 387–399.

Ratanamahatana, C. A., and Keogh, E. (2004). Everything you know about Dynamic Time Warping is Wrong. *Third Workshop on Mining Temporal and Sequential Data* (pp. 22-25).

Rath, T. M., and Manmatha, R. (2002). Lower-Bounding of Dynamic Time Warping Distances for Multivariate Time Series. Tech Report MM-40, University of Massachusetts Amherst.

Sakoe, H., and Chiba, S. (1978). Dynamic programming algorithm optimization for spoken word recognition. *Acoustics, Speech and Signal Processing, IEEE Transactions on 26*(1), 43–49.

Serra, J., and Arcos, J. L. (2012). A Competitive Measure to Assess the Similarity Between Two Time Series. In *Case-Based Reasoning Research and Development* (pp. 414–427). Springer Berlin Heidelberg.

Shabtai A., **Shahar Y**., and Elovici Y. (2006a). Using the knowledge-based temporal-abstraction (KBTA) method for detection of electronic threats. *Proceedings of The 2006 Fifth European Conference on Information Warfare and Security* (*ECIW*), Helsinky, Finland.

Shabtai, A., **Shahar, Y.,** Elovici, Y. (2006b). "Monitoring For Malware Using a Temporal-Abstraction Knowledge Base", *Proceedings of The 8th International Symposium on Systems and Information Security* (SSI 2006), Sao Jose dos Campos, Sao Paulo, Brazil.

Shabtai, A., Klimov, D., **Shahar, Y.,** and Elovici, Y. (2006). An Intelligent, Interactive Tool for Exploration and Visualization of Time-Oriented Security Data. Proceedings of *The Workshop of Visualization in Information*

*Security* as part of *The 13th ACM Conference on Computer and Communication Security* (*CCS 2006*), Alexandria, VA, USA, 2006.

Shabtai, A., Fledel, Y., Elovici, Y., and **Shahar, Y**. (2010). Using the KBTA method for inferring computer and network security alerts from time-stamped, raw system metrics. *Journal of Computer Virology and Hacking Techniques* (Formerly *Journal in Computer Virology*) **6** (3), 239-259.

Shahar, Y. (1997). A framework for knowledge-based temporal abstraction. *Artificial Intelligence* 90 (1–2), 79–133.

Shahar, Y. (1998). Dynamic temporal interpretation contexts for temporal abstraction. *Annals of Mathematics and Artificial Intelligence* 22 (1-2) 159-192.

Shahar, Y. (1999). Knowledge-based temporal interpolation. *Journal of Experimental and Theoretical Artificial Intelligence* 11, 123-144.

Shahar Y. and Combi C. (1998). Timing Is Everything: Time-Oriented Clinical Information Systems, *Western Journal of Medicine*, *168*(2), 105.

Shahar, Y., and Musen, M. A. (1993). RESUME: a temporal-abstraction system for patient monitoring. *Computers and Biomedical Research, an International Journal*, *26*(3), 255–273.

Shahar, Y., and Musen, M.A. (1996). Knowledge-based temporal abstraction in clinical domains. *Artificial Intelligence in Medicine* 8 (3), 267–298.

Shahar Y., Chen H., Stites D., Basso L., Kaizer H., Wilson, D., and Musen M.A. (1999). Semiautomated acquisition of clinical temporal-abstraction knowledge. *Journal of the American Medical Informatics Association* 6(6), 494-511.

Shahar, Y., Goren-Bar, D., Boaz, D., and Tahan, G. (2006). Distributed, intelligent, interactive visualization and exploration of time-oriented clinical data and their abstractions. *Artificial Intelligence in Medicine*, *38*(2), 115–135.

Sheetrit, E., Nissim, N., Klimov, D., and Shahar, Y. (2019). Temporal probabilistic profiles for sepsis prediction in the ICU. In: Proceedings of *The 25th ACM SIGKDD Conference on Knowledge Discovery and Data Mining* (*KDD-2019*), Anchorage, Alaska, the USA.

Shknevsky, A., Shahar, Y., and Moskovitch, R. (2017). Consistent discovery of frequent interval-based temporal patterns in chronic patients' data. *The Journal of Biomedical Informatics* 75:83-95.

Sun, R., and Giles, C. L. (2001). *Sequence Learning: Paradigm, Algorithm, and Applications* (Vol. 1828). Springer Science & Business Media.

Swan, M. (2009). Emerging patient-driven health care models: an examination of health social networks, consumer personalized medicine and quantified self-tracking. *International Journal of Environmental Research and Public Health*, *6*(2), 492–525.

Tormene, P., Giorgino, T., Quaglini, S., and Stefanelli, M. (2009). Matching incomplete time series with dynamic time warping: an algorithm and an application to post-stroke rehabilitation. *Artificial Intelligence in Medicine*, *45*(1), 11–34.

Vlachos, M., Hadjieleftheriou, M., Gunopulos, D., and Keogh, E. (2005). Indexing Multidimensional Time-Series. *The VLDB Journal—The International Journal on Very Large Data Bases*, *15*(1), 1–20.

Wang, X., Mueen, A., Ding, H., Trajcevski, G., Scheuermann, P., and Keogh, E. (2012). Experimental comparison of representation methods and distance measures for time series data. *Data Mining and Knowledge Discovery*, *26*(2), 275–309.

# APPENDIX

(See Section 2.1 for the semantics of the knowledge-based temporal-abstraction properties).

**Table A1** - The knowledge base for the **Oncology** domain. Each concept's knowledge includes the State abstractions numeric values range and the Gradient abstractions significant variations. GB = Good Before; GA = Good After

| White Blood-Cell (WBC) Count ($10^3/mm^3$) (Max (GB, GA) = 1 Day) || | PLATELET Count ($10^3/mm^3$) (Max (GB, GA) = 1 Day) || |
|---|---|---|---|---|---|
| State || Gradient | State || Gradient |
| State | Value Range | Significant Variation | State | Value Range | Significant Variation |
| VERY HIGH | ≥20 | 0.1 | HIGH | ≥400 | 20 |
| HIGH | 12-20 | | NORMAL | 100-400 | |
| NORMAL | 2.5-12 | | MODERATELY LOW | 50-100 | |
| MODERATELY LOW | 0.5-2.5 | | LOW | 20-50 | |
| LOW | 0.1-0.5 | | VERY LOW | <20 | |
| VERY LOW | <0.1 | | | | |
| Hemoglobin (HGB) Value (Gr/dL) (Max (GB, GA) = 1 Day) || | BAND Cells Count ($10^3/mm^3$) (Max (GB, GA) = 1 Day) || |
| State || Gradient | State || Gradient |
| State | Value Range | Significant Variation | State | Value Range | Significant Variation |
| HIGH | ≥16 | 0.8 | HIGH | ≥0.6 | 0.06 |
| NORMAL | 11-16 | | NORMAL | <0.6 | |
| MODERATELY LOW | 9-11 | | | | |
| LOW | 7-9 | | | | |
| VERY LOW | <7 | | | | |
| MONOCYTE Count ($10^3/mm^3$) (Max(GB, GA) = 1 Day) || | | | |
| State || Gradient | | | |
| State | Value Range | Significant Variation | | | |
| HIGH | ≥0.10 | 0.02 | | | |
| NORMAL | 0.3-0.10 | | | | |
| LOW | <0.3 | | | | |

**Table A2** - The knowledge base for the **Hepatitis** domain. Each concept's knowledge includes the State abstractions numeric values range and the Gradient abstractions significant variations (in relative or absolute terms).

| ALP (Units/L) (Max(GB, GA) = 3 Days) | | | LDH (Units/L) (Max(GB, GA) = 7 Days) | | |
|---|---|---|---|---|---|
| **State** | | **Gradient** | **State** | | **Gradient** |
| State | Value Range | Significant Variation | State | Value Range | Significant Variation |
| HIGH | >120 | 20% | HIGH | >280 | 20% |
| NORMAL | 30-120 | | NORMAL | 140-280 | |
| LOW | <30 | | LOW | <140 | |
| **Indirect BILIRUBIN (mg/dL) (Max(GB, GA) = 1 Day)** | | | **Total BILIRUBIN (mg/dL) (Max(GB, GA) = 1 Day)** | | |
| **State** | | **Gradient** | **State** | | **Gradient** |
| State | Value Range | Significant Variation | State | Value Range | Significant Variation |
| HIGH | > 0.9 | 20% | HIGH | >1.2 | 20% |
| NORMAL | 0.2-0.9 | | NORMAL | 0.2-1.2 | |
| LOW | <0.2 | | LOW | <0.2 | |
| **Direct BIL (mg/dL) (Max(GB, GA) = 1 Day)** | | | | | |
| **State** | | **Gradient** | | | |
| State | Value Range | Significant Variation | | | |
| HIGH | ≥0.3 | 20% | | | |
| NORMAL | <0.3 | | | | |

**Table A3** - The knowledge base for the **Diabetes** domain. Each concept's knowledge includes the State abstractions numeric values range and the Gradient abstractions significant variations.

| Albuminuria-U24h (mg/24h)<br>(Max(GB, GA) = 3 Months) | | | | | |
|---|---|---|---|---|---|
| FEMALE | | | MALE | | |
| State | | Gradient | State | | Gradient |
| State | Value Range | Significant Variation | States | Values Range | Significant Variation |
| MACRO | >300 | 20% | MACRO | >300 | 20% |
| MICRO | 30-300 | | MICRO | 30-300 | |
| NORMO-HIGH | 13-30 | | NORMO-HIGH | 15-30 | |
| NORMO-LOW | 0-13 | | NORMO-LOW | 0-15 | |
| **CREATININE (mg/dL)**<br>**(Max(GB, GA) = 2 Months)** | | | | | |
| FEMALE | | | MALE | | |
| State | | Gradient | State | | Gradient |
| State | Value Range | Significant Variation | State | Value Range | Significant Variation |
| VERY HIGH | >4 | 0.18 | VERY HIGH | >4 | 0.18 |
| HIGH | 2-4 | | HIGH | 2-4 | |
| MODERATELY-HIGH | 1.1-2 | | MODERATELY-HIGH | 1.3-2 | |
| NORMAL | <1.1 | | NORMAL | <1.3 | |
| **HbA1c (%)**<br>**(Max(GB, GA) = 6 Months)** | | | **CHLORIDE (mEq/L)**<br>**(Max(GB, GA) = 2 Months)** | | |
| State | | Gradient | State | | Gradient |
| State | Value Range | Significant Variation | State | Value Range | Significant Variation |
| VERY HIGH | >10.5 | 0.8 | VERY HIGH | >107 | 2 |
| HIGH | 9-10.5 | | HIGH | 106-107 | |
| MODERATELY-HIGH | 7-9 | | NORMAL | 98-105 | |
| NORMAL | <7 | | LOW | <98 | |

# Interval-Based Adjacent Proportional (IBAP) Interpolation

As mentioned briefly in the Methods section, with results presented in the Discussion Section, we also experimented, separately from our main evaluation, with a novel approach that we had developed that exploits, within the granularity-based representation step of the iDTW methodology, the previous step of abstracting the data into interval-based, knowledge-based temporal abstractions. We refer to this approach as the *Interval-Based Adjacent Proportional interpolation* (*IBAP*) method. IBAP is inspired by the method developed for visualization and exploration of patient records at different granularities, which was previously introduced by Klimov et al. [2010a].

Using the IBAP method, the values of a state-abstraction concept within a gap of any duration that exists between two intervals (also of any duration) that have *different* state-abstraction values for that concept are generated by considering the duration of the two intervals adjacent to the empty (i.e., without an assigned concept value) interval, and extending the values of the concepts of the two neighboring intervals into the empty time-interval gap in a manner proportional to the durations of the two neighboring intervals, such that the duration of each value within the gap is proportional to the duration of the respective adjacent intervals on both sides of the previously empty interval, over which each value holds. So if we have for two days a High value of the Hemoglobin-value state abstraction, followed by three empty days, followed by one day with a Low value of the Hemoglobin-value state abstraction, the first two days within the empty period will be assigned the value High, and the third day will be assigned the value Low.

(Note that if, at the end of the IBAP process, and after consideration had been given to all empty intervals bounded on both sides by intervals with some value for the concept at hand, empty time-granules exist at the edges of an entity's timeline, for some concept, their value will be induced by only one neighbor, their adjacent one. Typically, due to the scoping, this situation might be relevant only to the beginning of the timeline).

Note that all of the domain-specific interpolation that was possible using the temporal-abstraction KB, such as by using the half-life of the Hemoglobin state-abstraction values, as exploited within the temporal-abstraction interpolation function, was already performed previously during the temporal-abstraction process. (The KBTA method uses domain-specific properties such as the persistence of a concept into the past and into the future, using *local persistence functions* [Shahar, 1999]). Thus, the IBAP interpolation procedure is in a sense a *meta-interpolation* method, applied beyond the application of such temporal-interpolation methods, and designed mostly as an alternative option for enabling the DTW methods to work smoothly during the iMatch phase.

One further justification for the IBAP meta-interpolation procedure, beyond the fact that we wanted to "pad" the empty spaces in each patient's timeline in a semantically meaningful fashion, so as to enable the DTW procedure to work, is the fact that studies by Hripcsac et al. [2015] have demonstrated that a sequence-based parameterization of the timeline in medical records, which simply counts the number of measurements from some start, and sorts them in an ordinal (non-metric) fashion, best explained the variability in longitudinal-data studies. Sequence time appeared to produce the most stationary series, *possibly because clinicians adjust their sampling to the acuity of the patient* [Hripcsak et al., 2015]. Thus, during longer "empty" time periods, clinical time-series values tend to vary at a slower rate than when the values are sampled frequently, perhaps due to a clinically unstable condition.

Finally, consider also that most *global persistence functions*, namely functions which determine the maximal temporal gaps allowed between two intervals of the same value in order to fill in the gap using that value, are *positive-positive monotonic* in the case of the medical domain, namely, allow for a longer gap between intervals, the longer the duration of either the interval preceding the gap or the interval succeeding it [Shahar, 1999]. Thus, proportional "padding" seems a reasonable alternative to other, standard method.

In our separate experiment, which was *not* a part of testing our main research questions regarding the value of temporal abstractions versus the value of raw data, we examined the effect of IBAP as a new interpolation-function alternative to the three standard interpolation methods mentioned above. We briefly report the results in the Discussion, since they do not affect the answers to the primary or even the secondary research questions that we defined in Section 5.

# Constraining the Warping Window through a Knowledge-Based Band (KB-Band)

To avoid pathological matching, prevent irrelevant patterns from being discovered, and also to accelerate the matching process by narrowing the possible warping paths needed to be considered within the cost matrix, the Sakoe-Chiba band is usually used for constraining the warping window width [Sakoe and Chiba, 1978]. The typical values used in most studies are a window size whose duration is 10% of the overall time-series duration, or Infinity (no constraint). These are the two values we used in our main evaluation of the iDTW methodology.

We also experimented, completely separately from our main evaluation, as we briefly discuss in the Discussion Section, with a novel window-constraining approach that we have defined. Since the maximal window span of 10% of the overall time series duration seems rather arbitrary, as previously noted by several researchers [Ratanamahatana and Keogh, 2004], we have defined a new *Knowledge-Based-Band (KB-Band)*. Inspired by the Sakoe-Chiba band, the KB-Band implements a domain-specific temporal band, or window size, defined as the maximal half-life of all of the matched concepts $C$ involved in the matching task, to induce the value $r$ (the size of the band) of the Sakoe-Chiba band.

The physical or biological half-life of a substance, also denoted as $t_{\frac{1}{2}}$, defines the period of time that it takes the substance to lose half of its pharmacological, physiological, or radiological activity. Somewhat analogously, the KBTA method considers the temporal persistence into the future as well as into the past, of the truth value of a concept, and in particular, an abstract concept (e.g., "State of Hemoglobin is Moderate Anemia, in the context of a young woman"), using *Local persistence functions* [Shahar, 1999]. The equivalent to the half-life time in the case of persistence is the temporal distance from the original measurement, in which the probability that the concept's value is the same as measured or as inferred is 50%. In our framework, we represented the local persistence of the truth of a parameter instance, given its current value, and the current context, into the past, in a simplified fashion, as a *Good-Before* (*GB*) constant duration, and its persistence into the future as *Good-After* (*GA*) constant duration. The GB and GA values of each (abstract) concept are a part of the domain-specific KB. Typically they are assumed to be equal, as a default, and are acquired from a domain expert or an equivalent source.

Given a set of concepts $C$ participating in the matching process, the proposed *KB-Band* can be described by modification of the restriction range $r$ as follows:

$$r = Max\left(t_{\frac{1}{2}}^{c_1}, t_{\frac{1}{2}}^{c_2}, \ldots, t_{\frac{1}{2}}^{c_n}\right), n = |C|, c_i \in C$$

where

$$t_{\frac{1}{2}}^{c_i} = Max(GB_{c_i}, GA_{c_i})$$

In which $t_{\frac{1}{2}}^{c_i}$ refers to the half-life of concept $c_i$, $GB_{c_i}$ to the Good-Before persistence, and $GA_{c_i}$ to the Good-After persistence, of $c_i$.

In the separate experiment, we examined the value of using the KB-Band as an alternative to a 10% maximal-duration warping-window size, and to the use of no constraints. We briefly discuss the results of this experiment in the Discussion section, since they do not affect our main set of experiments.